\documentclass{article}

\usepackage[preprint,nonatbib]{neurips_2022}

\usepackage[utf8]{inputenc} % allow utf-8 input
\usepackage[T1]{fontenc}    % use 8-bit T1 fonts
\usepackage{hyperref}       % hyperlinks
\usepackage{url}            % simple URL typesetting
\usepackage{booktabs}       % professional-quality tables
\usepackage{amsfonts}       % blackboard math symbols
\usepackage{nicefrac}       % compact symbols for 1/2, etc.
\usepackage{microtype}      % microtypography
\usepackage{xcolor}         % colors
\usepackage{graphicx}
\usepackage{enumitem}
\usepackage{algorithm}
\usepackage{algorithmic}
\usepackage{makecell}
\usepackage{authblk}
\usepackage{bbold}
\usepackage{tablefootnote}
\usepackage{footnote}
\usepackage{amsmath}

\newcommand*\samethanks[1][\value{footnote}]{\footnotemark[#1]}

\title{Nebula-I: A General Framework for Collaboratively Training Deep Learning Models on Low-Bandwidth Cloud Clusters}

\author[1\thanks{Equal contributions.}]{Yang Xiang}
\author[2\samethanks]{Zhihua Wu}
\author[2\samethanks]{Weibao Gong}
\author[2\samethanks]{Siyu Ding}
\author[3,4,1\samethanks]{Xianjie Mo}
\author[2]{Yuang Liu}
\author[2]{Shuohuan Wang}
\author[2]{Peng Liu}
\author[1]{Yongshuai Hou}
\author[2]{Long Li}
\author[2]{Bin Wang}
\author[5]{Shaohuai Shi}
\author[2]{Yaqian Han}
\author[1]{Yue Yu}
\author[1]{Ge Li}
\author[2]{Yu Sun}
\author[2]{Yanjun Ma}
\author[2]{Dianhai Yu}
\affil[1]{
Peng Cheng Laboratory, Shenzhen, China. \texttt{\{xiangy, moxj\}@pcl.ac.cn}
} 
\affil[2]{
Baidu Inc., Beijing, China. \texttt{\{wuzhihua02, gongweibao, dingsiyu\}@baidu.com}
}
\affil[3]{
Key Lab of Intelligent Information Processing of Chinese Academy of Sciences (CAS), Institute of Computing Technology, CAS, Beijing, China.
}
\affil[4]{
University of Chinese Academy of Sciences, Beijing, China.
}
\affil[5]{Hong Kong University of Science and Technology, Hong Kong, China.}
\begin{document}

\maketitle

\begin{abstract}
  The ever-growing model size and scale of compute have attracted increasing interests in training deep learning models over multiple nodes. However, when it comes to training on cloud clusters, especially across remote clusters, huge challenges are faced. In this work, we introduce a general framework, Nebula-I, for collaboratively training deep learning models over remote heterogeneous clusters, the connections between which are low-bandwidth wide area networks (WANs). We took natural language processing (NLP) as an example to show how Nebula-I works in different training phases that include: a) pre-training a multilingual language model using two remote clusters; and b) fine-tuning a machine translation model using knowledge distilled from pre-trained models, which run through the most popular paradigm of recent deep learning. To balance the accuracy and communication efficiency, in Nebula-I, parameter-efficient training strategies, hybrid parallel computing methods and adaptive communication acceleration techniques are jointly applied. Meanwhile, security strategies are employed to guarantee the safety, reliability and privacy in intra-cluster computation and inter-cluster communication. Nebula-I is implemented with the PaddlePaddle deep learning framework, which can support collaborative training over heterogeneous hardware, e.g. GPU and NPU. Experiments demonstrate that the proposed framework could substantially maximize the training efficiency while preserving satisfactory NLP performance. By using Nebula-I, users can run large-scale training tasks over cloud clusters with minimum developments, and the utility of existed large pre-trained models could be further promoted. We also introduced new state-of-the-art results on cross-lingual natural language inference tasks, which are generated based upon a novel learning framework and Nebula-I.
\end{abstract}

\section{Introduction}

The rapid evolution of pre-trained models are toward the trend of involving more and higher-quality data, larger amount of parameters and stronger computing power \cite{BERT,GPT-2,GPT-3,CPM-2,PANGU,ERNIE-Titan,Gopher,Switch,M6,PaLM}. It naturally relies on increasingly more compute cost and time, e.g. training the BERT-large 345 million model took 6.16E PF-days, training the GPT-3 175 billion model consumed 3.64E+03E PF-days \cite{GPT-3}. On the one hand, larger models tend to obtain better performances, especially on few- and zero-shot learning tasks \cite{ERNIE-Titan, PaLM}, which can greatly empower the AI industry. On the other hand, the increasing demand of compute brings about challenges and triggers more explorations on the development of advanced distributed training techniques as well as the optimization of large-scale resource scheduling and allocation strategies.

%e.g. training the BERT-large 345 million model took 6.16E PF-days, training the GPT-3 175 billion model consumed 3.64E+03E PF-days of compute and it reportedly cost \$12 Million for a single training run using a Tesla V100 cloud instance, and the recent PaLM even used up to 6,144 TPU chips and may cost around \$9\--17 Million \cite{big-model-cost}

%discuss parallel computing%
There have been extensive researches that focus on parallel computing for training large deep learning models. The most popular techniques include data parallism \cite{PS, Poseidon}, model parallism \cite{Mesh, Megatron}, pipeline parallism \cite{Pipedream, Memory-efficient-pp} and hybrid parallism \cite{Megatron, DeepSpeed}, which aim to accelerate the training process from different dimensions. These techniques have already been widely applied in the training of recent pre-trained models and obtained considerable speed-up ratios \cite{GPT-3, PANGU, ERNIE-Titan, PaLM}.

%transfer to cloud-based parallellization%
As the model size continues to grow, e.g. from 170 billion \cite{GPT-3} to 540 billion \cite{PaLM}, more problems arise. One of the practical problems is: it will become more frequent that the spare resources one cluster can provide are insufficient to satisfy the training of a large model, which is particular common for business clusters. A direct intuition to address the problem is to expand the computing power by building larger-scale of hardware. With the development of nationwide infrastructure construction, many organizations have built super computers or intelligent clusters distributed in different locations. Each of these clusters, however, only serves independently in most cases. Shi et al. introduced a method to train deep learning models on public cloud clusters, which utilizes distributed compute connected by networks \cite{Public-Cloud}. This inspires us to explore potential solutions to connect the computational resources of different cloud clusters and train models in a collaborative style. Meanwhile, the modularity of many deep learning algorithms enables the decoupling of a model into different components, which could be dispatched to multiple clusters. However, there also exist many challenges. For example, as the connections between cloud clusters are usually low-bandwidth (even much lower than those mentioned in \cite{Public-Cloud}), especially between different locations, communication efficiency will easily become a bottleneck.

To take a step forward, in this work, we proposed Nebula-I, a unified framework for collaboratively training deep learning models over public cloud clusters, especially on clusters connected by low-bandwidth wide area networks (WANs). Nebula-I is a stack of optimization layers together with a security layer, which is designed to better assist the deployment of a deep learning model on the cloud environment. Using Nebula-I, a model can be decoulpled and dispatched to different clusters which are then work collaboratively to execute certain tasks. Our motivations of proposing Nebula-I are in three folds. Firstly, we aim to set up practical solutions for pre-training larger models collaboratively over cloud clusters, in which the tasks are carried out using compute aggregated from different locations. Secondly, we would like to explore more possibilities that the cloud environment can support, to fully realize the functionality of cloud clusters and existed pre-trained models. Lastly, we expect Nebula-I can provide user-friendly interfaces that can minimize the development of end users who are willing to train models over the clouds. 

%Remove some of the following parts to Method%

We demonstrated how Nebula-I works using two natural language processing (NLP) scenarios that include: a) pre-training a multilingual language model using two remote clusters, and b) fine-tuning a machine translation model using knowledge distilled from pre-trained models. These two steps form into the most popular training paradigm of recent deep learning methods, which can cover a large variety of training tasks. In our demonstration, we deployed Nebula-I on remote cloud clusters with heterogeneous architectures, i.e. GPU and NPU clusters from Baidu and Peng Cheng Laboratory. The connections between the clusters are low-bandwidth WANs (only up to 170Mbit/s and 60Mbit/s), which is significantly lower than those within a cluster. This raises great challenges for the communication efficiency especially for training cases that need frequent data exchanges.

%a GPU cluster from Baidu Cloud, and a GPU and an NPU cluster from Peng Cheng Cloud Brain (I and II).%

Experiments were conducted on both the scenarios to validate the effectiveness of Nebula-I. For Scenario-I, we show the computation efficiency using Nebula-I in pre-training over two remote clusters by comparing the throughput and verify that the model can converge well on the cloud environment. Specifically, we generated new state-of-the-art (SoTA) performance on a series of cross-lingual natural language inference tasks with the multilingual model pre-trained under a novel learning framework and Nebula-I. For Scenario-II, we compared the knowledge-enhanced model with Transformer-base \cite{Transformer} and show that the accelerated model can well preserve the performance running across clusters. Different communication optimization combinations have been compared as well. 

The contribution of this work can be summarized as follows:
\begin{itemize}[leftmargin=*]
\item We introduce a novel framework that well fits distributed deep learning over remote cloud clusters, which are connected by low-bandwidth networks;
\item We propose a unified optimization technique in Nebula-I, named Nebula-Optimizer, that can jointly optimize the training strategy, parallization and communication;
\item Both pre-training and fine-tuning scenarios have been validated on the cloud environment, demonstrating the effectiveness of the proposed framework;
\item We output a multilingual pre-trained language model, ERNIE-M Extra, which has obtained new SoTA results on cross-lingual natural language inference tasks, under the proposed novel multilingual learning framework and Nebula-I. 
\item To our knowledge, this is the first work that decently introduces collaborative training techniques over cloud clusters, which could inspire further researches to delve into this area.
\end{itemize}

\section{The Nebula-I framework} %

\begin{figure}
  \centering
  \includegraphics[scale=0.46]{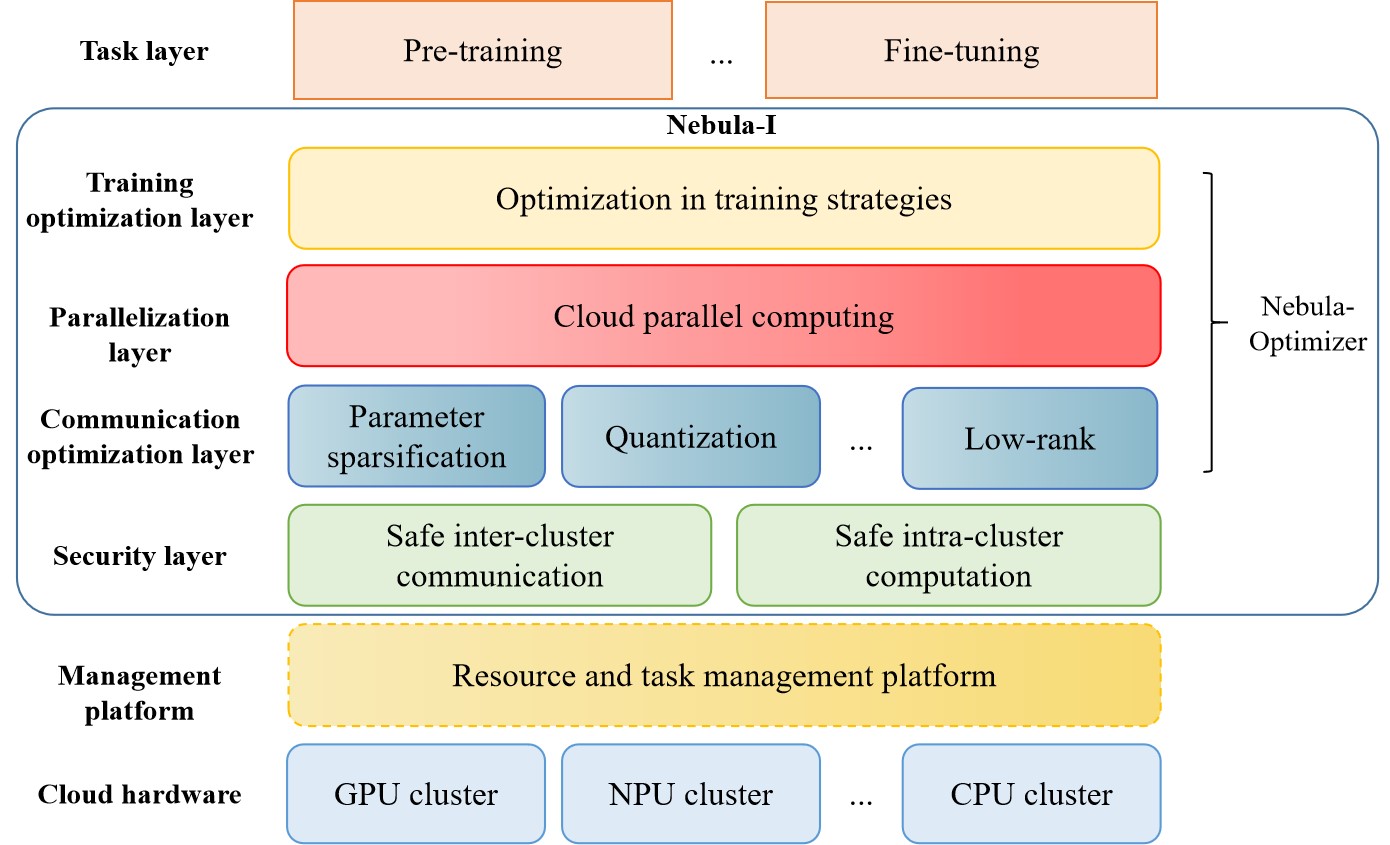}
  \caption{The general framework of Nebula-I. Note that we do not focus much on the management platform in this work (the dotted box).}
  \label{Nebula-I}
\end{figure}

The overview of Nebula-I is shown in Figure \ref{Nebula-I}. Nebula-I can be deployed on top of the cloud hardware and the management platform, and supports the user-specific tasks in the task layer. The cloud hardware layer is a group of machines that are connected using WANs and the hardwares can be heterogeneous. The hardwares distributed in different locations are managed using a unified management platform which can schedule and allocate resources in an automatic style (the management platform). After the computational resources are allocated and the training data are ready, users are able to run optimized distributed learning programs with the help of Nebula-I. The Nebula-I framework generally contains four layers:

\begin{itemize}[leftmargin=*]
\item {\it Training optimization layer}  The goal of
this layer is to provide communication-efficient techniques associated with user-specific deep learning tasks. Once a deep learning task, e.g. pre-train a language model, is selected by the user, this layer offers an entrance to define cloud-based training strategies, decouple the selected model and dispatch it to multiple clusters.

\item {\it Parallelization layer} This layer is to improve the computation efficiency through optimized scheduling of different tasks or operations. By adaptively applying the PaddlePaddle hybrid parallism techniques within each cluster, and designing task-specific distributed learning strategies across clusters, the training components dispatched to different clusters will work in parallel to execute the task collaboratively. When designing the parallization strategies, both the computation operations and the network environment should be taken into account, so as to maximize the overall throughput.

\item {\it Communication optimization layer} This layer includes an ensemble of data compression methods to accelerate the communication between clusters. For example, sparsification can be used to select the most effective set of elements during data transfer \cite{topk}. Quantization can further reduce the communication traffic by using low-bit numbers instead of the original data \cite{qsgd}. Singular value decomposition is a typical method in low-rank to select the most important feature representations \cite{Compression-survey}.

\item {\it Security layer} This layer offers security mechanisms for computation and communication for the whole training process across clusters. Through this layer, data confidentiality and security are ensured when collaboratively training models.
\end{itemize}

We argue that the execution of a model should be optimized from multiple perspectives before running on the cloud clusters. In Nebula-I, we call the joint of the three optimization layers (i.e. training strategy optimization, parallization, and communication optimization) \textit{Nebula-Optimizer} (Figure \ref{Nebula-I}), which forms a pipeline optimization system. Only by designing specific optimization strategies for each layer and let them work together can the overall running efficiency be maximized.

%  Different from previous studies, which mostly focus on reducing the communication traffic, we concentrate more on a pipeline optimization mechanism, i.e. each layer in Nebula-Optimizer can work jointly to achieve the overall compression and acceleration. 

% Nebula-Optimizer offers flexible interfaces for optimizing the model training over cloud clusters and is the core of the Nebula-I framework. We will provide user-friendly configurations for optimization on each layer and the selected sub-optimizers will work jointly to optimize the learning steps. Below we detailedly introduce how each component works in Nebula-Optimizer and how they can scale out in this subsection.

%  We will show how Nebula-I works below using a system that deployed on Baidu Cloud and Peng Cheng Cloud Brain clusters.

\subsection{Training optimization layer}
A targeted training strategy for cloud-based computing should satisfy both well convergence and reduced communication. There are several scenarios for cloud-based computing. For example, different clusters can be assembled in a parameter server (PS)-worker topology, where each worker computes multiple steps first and then the gradients are aggregated by the server. This PS-worker architecture works often for data parallism, where each worker owns its data and does not need to share with others. As each worker has the overview of a whole model, the volume for transfer could be huge, i.e. the size of the model, if the parameters between them are frequently exchanged, which could easily exhaust the network\cite{Poseidon}. On the other hand, if each worker is allowed to compute for sufficient local steps before parameter aggregation, the model convergence problem might exist \cite{FL-outlook}. Thus, when deploying Nebula-I on a PS environment, the communication frequency should be defined in advance, which might be designed together with the model's learning objective.

Another family of scenarios is where a complex model can be decoupled into multiple parts, and these parts can be located in different clusters. By applying pipeline parallism \cite{Pipedream}, the communication volume between clusters can be greatly reduced compared with data parallism, as generally only a proportion of parameters need to be transferred. However, the communication frequency is usually high, which can also affect the training efficiency. In the demonstration of the current work, we used this method as the overall architecture. However, we can scale out to more mechanisms, e.g. combine the above two together.

\begin{figure}
  \centering
  \includegraphics[scale=0.6]{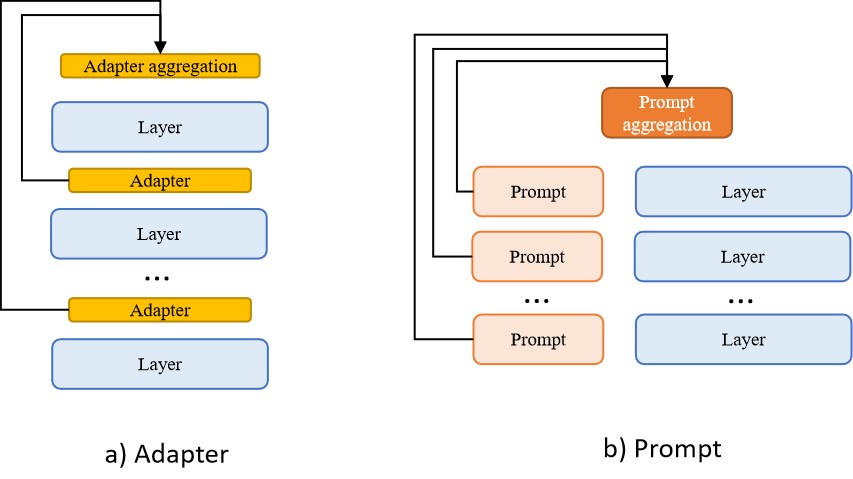}
  \caption{Popular light-weight training strategies.}
  \label{Cloud-service}
\end{figure}

Recently, some knowledge distillation (KD)-based studies offer new insights for training a model with multiple workers collaboratively \cite{ELECTRA, KD-MT}. We absorb these ideas into Nebula-I and argue that parameter-efficient techniques should be favored when a model is trained over clouds. In the current version of Nebula, we focus more on the support of KD-based pre-training and fine-tuning for the training of deep learning models. These models are usually easy to be decoupled into different parts, and require much fewer data to communicate compared with traditional training paradigms. We show two examples in Figure \ref{Cloud-service} where lightweight networks are added into existed large networks and only the parameters of the add-in networks are to be transferred between clusters \cite{ABNet, Prompt-MT}.

\subsection{Parallization layer}
Using PS is a direct way for distributed training especially data parallelism \cite{PS,Poseidon}. However, when frequent data exchange is met,  PS might not be a good choice as it can be overwhelmed by the huge volume of communications \cite{Geeps}, as mentioned above. Later work focus more on all-reduce together with other MPI primitives to form different types of parallization techniques including model parallism \cite{Mesh, Megatron}, pipeline parallism \cite{Pipedream, Memory-efficient-pp}, and hybrid parallism \cite{Megatron, DeepSpeed, PANGU, ERNIE-Titan}. 

The development of distributed training techniques has substantially accelerated the pace of training larger models \cite{Megatron, Pathways}. In Nebula-I, the training environment contains two parts, i.e. intra-cluster and inter-cluster. For the intra-cluster part, existing well-designed parallization techniques such as Megatron \cite{Megatron} and DeepSpeed \cite{DeepSpeed} can still be applied. For the inter-cluster part, as the bandwidth limitation exists, specific parallization techniques should be designed to consider both the interactions on computation between different clouds and the overlap between computation and network transfer. In Nebula-I, we have already supported data parallism, pipeline parallism, and model parallism. These techniques can also be assembled to form hybrid parallism. In our following demonstrations that show how Nebula-I works, the overall architecture of parallization for inter-cluster is pipeline parallism.

\subsection{Communication optimization layer}
Due to the low-bandwidth and high-latency network between different cloud clusters, data communication would easily become the system bottleneck. To address the communication problem in distributed training cross clusters, we integrate several communication optimization techniques in Nebula-Optimizer. In distributed training, data compression technologies including gradient or model quantization~\cite{qsgd}, sparsification~\cite{topk}, and low-rank updates~\cite{atomo, powersgd} are popular strategies to reduce the communication overheads\cite{Compression-survey}. 

Gradient or model quantization can exploit low-bit floating point number to represent the data that should be communicated through the network. For example, 8-bit quantization~\cite{8bit} only use 8-bit floating point to represent each gradient, which reduces 75\% communication traffic compared to the 32-bit counterpart while preserving the training convergence. Low-rank reduces the number of variables for communication by factorizing the source data matrix into several smaller matrices\cite{lowranksurvey}.
% reduces the number of bits for each variable to send by mapping data to less bits (limited-bit) or recoding the data to send (codebook-based). Converting data type from float 32 to float 16 is a simple limited-bit method which can reduce 50\% bits to send. 8-bit quantization\cite{8bit} maps float 32 to 8 bits, which reduces the source volume to 25\%. 

%  Sparsification reduces the number of variables to send by selecting a subset of variables from the source data. This method needs a bit map to indicate the position of each variable, and a strategy such as Random\mbox{-}\emph{k}\cite{randomk}, Top\mbox{-}\emph{k}\cite{topk}, Threshold\mbox{-}\emph{k}\cite{thresholdv}, to select variables to send. Suppose $n$ is the total number of variables in the source data, and $k$ is the number of variables to send, and the data type is 32-bit floating point, the compression ratio $R_s$ is calculated by Eq \ref{spars compress ratio}
 
% \begin{equation}
% R_s=\frac{k\times4\times8+n}{n\times4\times8}=\frac{k}{n}+\frac{1}{32}\label{spars compress ratio}
% \end{equation}

In Nebula-I, we use singular value decomposition (SVD) to factorize and decompose the data matrix $A\in\mathbb{R}^{m\times n}$ to three matrices by the following formula:
\begin{equation}
A=U\times S\times V^{T}
\end{equation}
where $U\in\mathbb{R}^{m \times m}$, $S\in\mathbb{R}^{m \times n}$, and $V\in\mathbb{R}^{n \times n}$. By using the low-rank feature of the matrix $A$, we only use some high singular values to compress the matrices while reserving the important information of the original matrix. Using
$U_r$, $S_r$ and $V_r$ to denote the compressed matrices for communication, they can be computed by
\begin{equation}
U_r=U[m,0:r], \quad
S_r=S[0:r,0:r],  \quad
V_r=V[n,0:r],
\end{equation}
where $r<n$. $r$ can be a hyper-parameter to tune the training performance while preserving the model accuracy. The data matrix can be recovered by the following formula:
\begin{equation}
A'=U_r\times S_r\times V_r^{T}
\end{equation}
where $A'$ is approximately equal to A. Thus, the compression ratio $R_{svd}$ using SVD is
\begin{equation}
R_{svd}=\frac{m\times r+r+r\times n}{m\times n}
\end{equation}

To further reduce the communication volume, we can quantize the compressed matrices from SVD compression with low-bit representation such as 16-bit floating points (FP16) or even 8-bit integers (INT8).
For example, for the input data $X_{input}$, we can compress the data by
\begin{equation}
X_{compressed}=C_{FP16}(C_{SVD}(X_{input}, r)),
\end{equation}
where $C_{FP16}$ is the quantization compressor that converts the 32-bit floating point numbers to 16-bit, $r$ is the number of singular values used.
% \begin{equation}
% X_{decompressed}=D_{FP16}(D_{SVD}(X_{compressed}))
% \end{equation}

% where $C_{SVD}$ and $D_{SVD}$ stand for SVD compression and decompression, $C_{FP16}$ is the float 32 to float 16 conversion function and $D_{FP16}$ is the float 16 to float 32 conversion.

\subsection{Security layer}
Compared with training models within a cluster, cross-cluster training brings more challenges to data security. This is mainly because the model parameters (e.g., weights or gradients) transferred between two clusters may leak sensitive information to malicious adversaries and cause deep privacy leakage \cite{fl-security1}. As reported in \cite{fl-security2}, a small portion of the original gradients could reveal privacy about local training datasets. Therefore, security layer in Nebula-I provides four mechanisms for computation and communication safety.

\begin{figure}[H]
  \centering
  \includegraphics[scale=0.08]{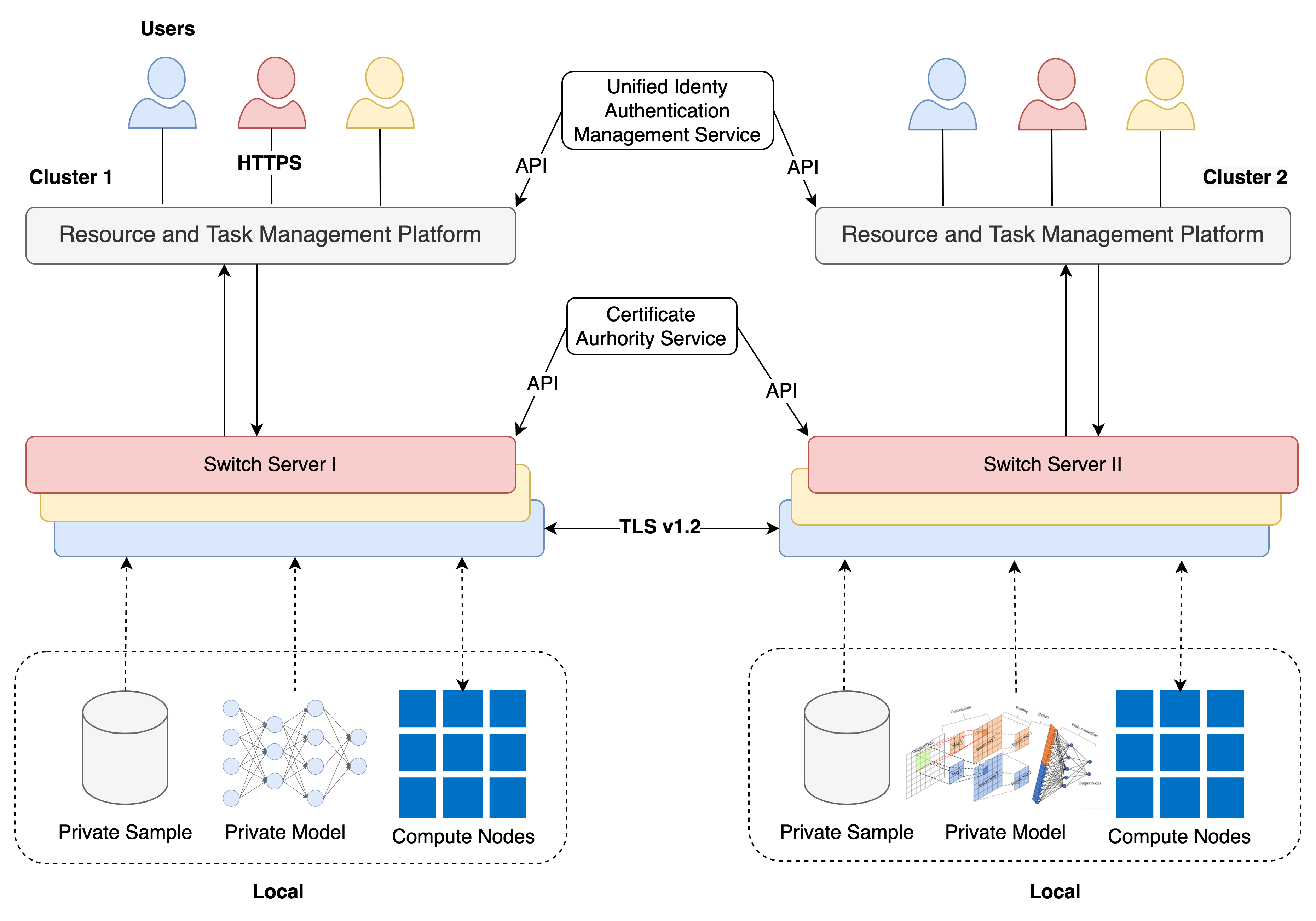}
  \caption{Computation and communication safety framework for cross-cluster training.}
  \label{switch}
\end{figure}

The first mechanism is to separate the computing nodes from the public network, e.g., Internet, while only one node (called "Switch Server") can access to outsiders. 

The second mechanism is a unified identity authentication management service, including account life cycle, uniformed data access control. When a new user needs to submit a task (e.g. upload data \& code), we create a unique account and a password. Then isolated resources (compute, storage and network) are allocated for each user. When a user applies for the permission of the data owned by another user, a data access interface is provided by the data owner instead of direct raw data sharing.

The third mechanism is encrypted data transfer using TLSv1.2 protocol across two clusters. A cloud certificate manager service is to issue certificates and manages the life cycle of certificates by calling APIs provided by cloud service providers. Switch Server in the local cluster applies for digital certificates with its own public key, ID, information of the issuing certificate authority (CA), validity time, certificate serial number and other information, including a signature. Meanwhile, a CA public key is sent to the peer Switch Server (as a client).

The last mechanism is the audition of the codes, data and operations, which ensures that no data is accessed and no action is performed illegally.

Through these mechanisms, data confidentiality and security are ensured when collaboratively training models over remote heterogeneous clouds.

\section{How Nebula-I works for pre-training and fine-tuning}
We took two clusters as our cloud environment to deploy Nebula-I in each scenario that we would like to demonstrate. The two scenarios are designed to show how Nebula-I works in a pre-training - fine-tuning process, which is a general pipeline of training deep learning models. In each scenario, we followed a top-down style of Nebula-I (Figure  \ref{Nebula-I}) to design the functionality of each layer and finally make them work together.

\subsection{Pre-training: ERNIE-M Extra-Cloud} 
The first scenario (Scenario-I) is designed to pre-train a new NLP model from an existed well-trained model, the design of which aims to maximize the utility of existed models but only use knowledge distilled from the cloud. This could be extremely useful when we do not have direct access to certain existed models, or we do not have enough capacity to maintain two model during training. Another, the collaborative training on cloud clusters makes it possible for collaborative training with multi-sources pretraining data. In this paper, we simulate the collaborative training on multi-sources pretraining data with multilingual pretraining. For training optimization, we employed the pre-training structure from ELECTRA \cite{ELECTRA}, the architecture of which contains a generator and a discriminator. The output of Scenario-I is a much larger multilingual language model named ERNIE-M Extra (550 Million parameters, defined as discriminator), distilled from a smaller ERNIE-M model \cite{ERNIE-M} (220 Million parameters, defined as generator). Notably, the prevailing ELECTRA-based pre-trained models \cite{ELECTRA, XLM-E} advocated sharing the embeddings (both the token and positional embeddings) between the generator and discriminator to improve the efficiency of pre-training, this undoubtedly increases the traffic between the two clusters. Furthermore, the shared embeddings forces the generator and discriminator to have the same hidden size, which severely limits the flexibility of Scenario-I. Consequently, we proposed to discard the strategy of sharing embeddings to minimize the communication between two clusters to only a small number of word indices. 

Different from the previous BERT-style pre-trained language models, that trained by predicting the masked tokens, ELECTRA innovatively proposed a sample-efficient pre-training task known as Replaced Token Detection (RTD) to train $discriminator$ to distinguish real input tokens vs fake input tokens generated by $generator$. Specifically, the generator $G$ is a conventional BERT-style model, which was trained with the Masked Language Modeling (MLM) \cite{BERT} task. Given the input $x=[x_1, x_2, ..., x_n]$ that consists of $n$ tokens, MLM replaces the tokens in a randomly selected position subset ${\mathcal{M}=\{m_1, m_2, ..., m_m~|~m<n\}}$ with \texttt{[MASK]} token to construct the masked input $x^{\texttt{mask}}$. Then, the generator learns to predict the probability distributions of the masked-out tokens $p_G(x_i | x^{\texttt{mask}})$. The discriminator $D$ is another Transformer-based encoder, trained with the RTD task, in which the corrupted input $x^\texttt{corrupt}$ is built by displacing the tokens in $\mathcal{M}$ with the generated sample. Formally, the loss function $\mathcal{L}_{E}$ of ELECTRA is as follows:
\begin{small}

\begin{equation}
\mathcal{L}_{G}(x;\theta_{G})=-\sum_{i \in \mathcal{M}}log~p_{G}(x_i|x^{\texttt{mask}}),
\end{equation}

\begin{equation}
\left \{
    \begin{array}{ll}
        x_{i}^{corrupt} = p_G(x_i | x^{\texttt{mask}}),  & i \in \mathcal{M}\\
        x_{i}^{corrupt} = x_i,     & i \notin \mathcal{M},
    \end{array}
\right.
\end{equation}

\begin{equation}
\mathcal{L}_{D}(x;\theta_{D})=-\sum_{i=1}^{n}\mathbb{1}(x_{i}^{corrupt} = x_i)log~D(x^{corrupt}, i)+\mathbb{1}(x_{i}^{corrupt} \ne x_i)log~(1 - D(x^{corrupt}, i)),
\end{equation}

\begin{equation}
\mathcal{L}_{E} = \mathcal{L}_{G} + \lambda\mathcal{L}_{D}
\end{equation}

\end{small}

\begin{figure}
  \centering
  \includegraphics[scale=0.38]{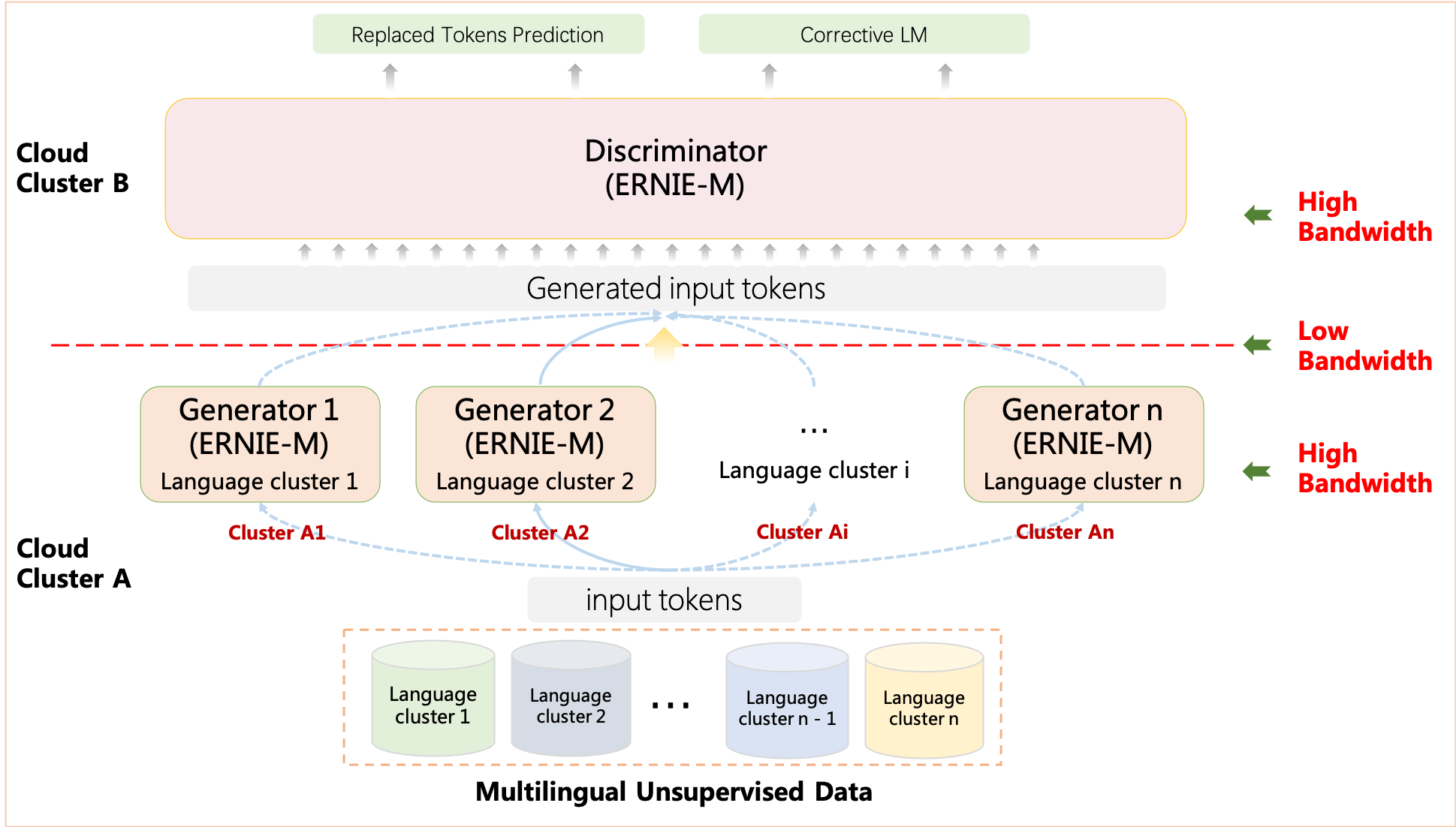}
  \caption{The framework of ERNIE-M Extra-Cloud.}
  \label{scneI-ernie-m-cloud}
\end{figure}

\paragraph{Framework}
Inspired by the language grouping \cite{M2M-100} among multilingual and the collaborative training on cloud clusters, we creatively propose the framework of \textit{\textbf{M}ulti-sources \textbf{M}ulti-lingual pre-training with \textbf{M}ulti-tasks} (\textbf{M3}). As shown in Figure \ref{scneI-ernie-m-cloud}, the M3 framework constructs n generators for the n language clusters grouped by language family that is introduced in \cite{M2M-100}. Evidently, a large multilingual corpus with more languages requires models with increased capacity, expanding the scale of the model is a common-used strategy. Under the Scenario-I, we explore to employ more generators for different language clusters to learn the corresponding language knowledge, which lead to four benefits, that is, (1) language cluster with fewer languages demands weaker model capability (such as the capability of base model size); (2) the pre-training strategies of different language clusters trained on different generators are more beneficial to the knowledge transfer between similar languages; (3) the multi-sources (generators) structure fits perfectly with the collaboratively training framework on multi cloud clusters (each generator can be deployed on one cluster); (4) different pre-training tasks can be conducted on different generators, which resulting multi-task learning. For the discriminator, the multi-sources (generator) can be regarded as multi-teachers and the fusion knowledge from different teachers enables the efficient training of discriminator.

\paragraph{Pre-training Tasks}
Following XLM-E \cite{XLM-E} and COCO-LM \cite{COCO-LM}, we construct three tasks, relied on self-supervised or weak-supervised
signals that could be obtained from massive data without human annotation, to pre-train ERNIE-M Extra, that is  Multilingual Replaced Token Detection (MRTD) task, Translation Replaced Token Detection (TRTD) task and Corrective Language Modeling (CLM) task.
\begin{itemize}
\item \textbf{Multilingual Replaced Token Detection} As introduced in \cite{XLM-E}, the multilingual replaced token detection task is similar to that in monolingual ELECTRA pretraining, of which input sentences can spread across various languages rather than a single language. Naturally, the generator, the discriminator and the vocabulary are all shared across languages.

\item \textbf{Translation Replaced Token Detection} As the opening explore for how to improve discriminative pre-training based on parallel corpora, translation replaced token detection, proposed in \cite{XLM-E}, aims to distinguish real input tokens from corrupted parallel sentence pairs. At length, consider an input parallel sentence pair, MLM first chooses a random positions subset to be masked, in which the positions are uniformly distributed in both languages. The Generators learn to predict the masked tokens and then construct the corrupted parallel sentence pairs by replacing the masked-out tokens with generated samples. The discriminative pre-training is conducted on discriminator through predicting whether the token corrupted parallel sentence pairs is the original one or the replaced one. 

\item \textbf{Corrective Language Modeling} Compared to BERT-style pre-trained models, ELECTRA is more compute-efficient and achieves better performance. While the lack of language modeling ability limits the application of the model, mentioned in \cite{COCO-LM}. COCO-LM proposed the corrective language modeling task to alleviate the problem described above. More specifically, given the corrupted input $x^\texttt{corrupt}$, CLM trains the discriminator to recover the original tokens by optimizing RTD task and All-Token MLM task (the detail describe can refer to \cite{ELECTRA}) simultaneously. Nevertheless, the RTD loss has been calculated in both multilingual replaced token detection task and translation replaced token detection for ERNIE-M Extra. Therefore, the loss function of corrective language modeling is simplified as follows,
\begin{small}
\begin{equation}
    \mathcal{L}_{CLM} = \sum_{i \in \mathcal{M}} p_{LM} (x_i|h_i)
\end{equation}
\end{small}

\end{itemize}

In general, we minimize the combined loss
\begin{small}
\begin{equation}
    min_{\theta_{G_1}, ..., \theta_{G_K}, \theta_{D}}\sum_{x \in \mathcal{X}} \mathcal{L}_{G} + \lambda \mathcal{L}_{D} + \gamma \mathcal{L}_{CLM}
\end{equation}
\end{small}
over a large monolingual and parallel corpora $\mathcal{X}$, in which the $G_{i}(i \in \{1..K\})$ denotes the i-th generator deployed on the corresponding i-th cloud cluster. Similar to ELECTRA, only the discriminator is fine-tuned on downstream tasks while the generators are discarded.

\subsection{Fine-tuning: ABNet-Cloud}
\label{section:Fine-tuning-ABNet-Cloud}

\begin{figure}
  \centering
  \includegraphics[scale=0.8]{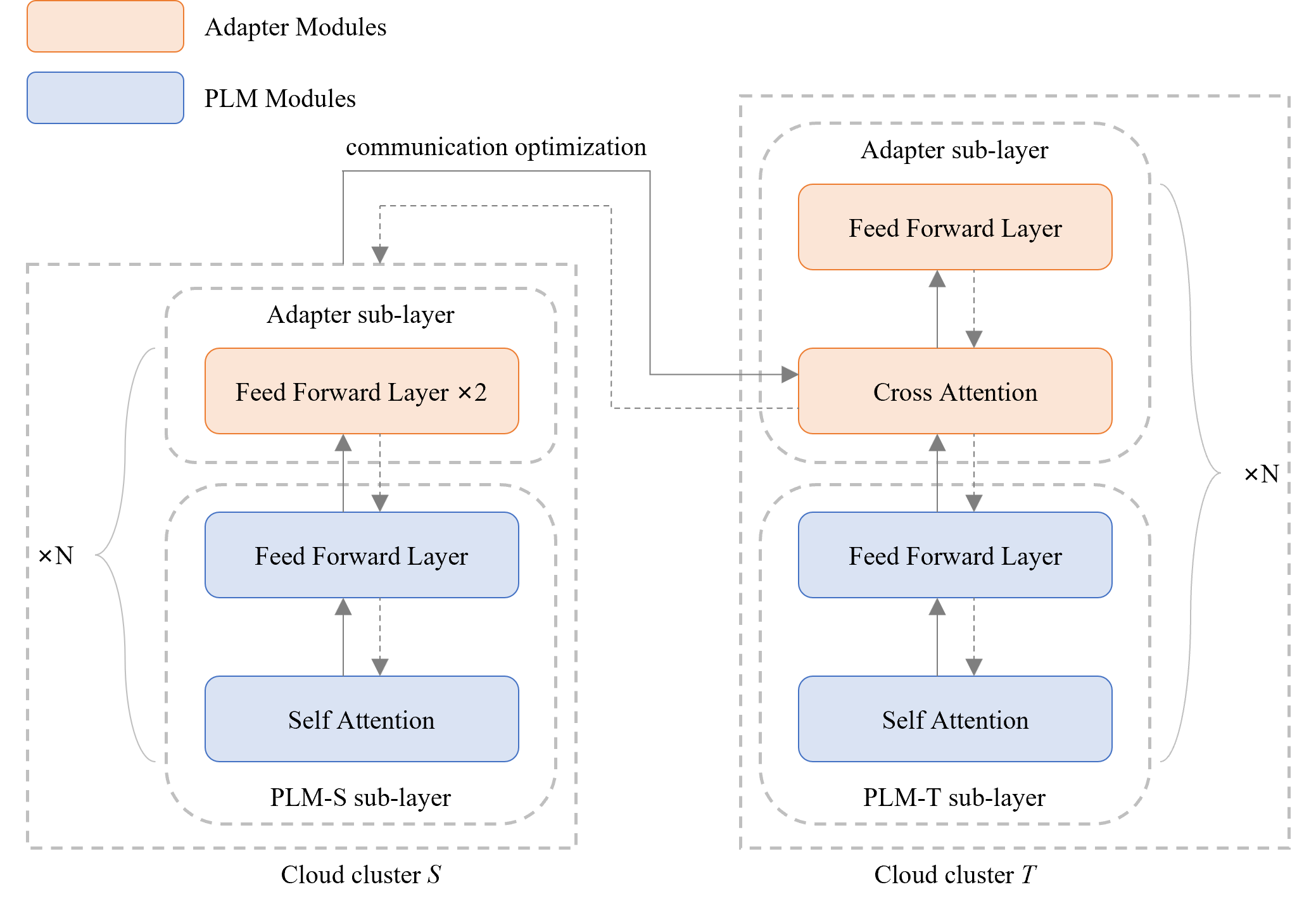}
  \caption{The framework of ABNet-cloud.}
  \label{scneII-abnet-cloud}
\end{figure}

The goal of the second scenario (Scenario-II) is to fine-tune a task model with the help of existed pre-trained models, e.g. ERNIE-M Extra from Scenario-I. We use machine translation (MT) as the fine-tuning task in this scenario. Specifically, we assume two pre-trained models for the source and target languages are distributed in two remote cloud clusters, respectively. During training the MT model, the task executor will iteratively require knowledge from the pre-trained models for encoding and decoding. We adopted the architecture of ABNet \cite{ABNet} to simulate this process, in which the encoder and decoder parts are initialized from two pre-trained models. Since the source and target models are decoupled, neither of the cloud clusters can obtain any parallel training instances, thus the privacy of the user's corpus can be guaranteed. 
%In both the scenarios, a user only needs to submit a task to the cloud computing platform, and the platform will assign the task to different clusters for execution and then return results to the user.%

ABNet is a parameter-efficient fine-tuning model which comprises of adapter modules and pre-trained models for the source and target languages. It utilizes knowledge from separated pre-trained models by training the inserted adapter modules between en/decoder sub-layers while freezing the parameters of pre-trained models. For the encoder side, the adapter module with layer normalization as well as two feed-forward layers with non-linear activation is inserted between each sub-layer of the pre-trained (masked) language model from the source language domain. As for the decoder side, the base module is the pre-trained (masked) language model from the target language domain, and the adapter module that consists of the multi-head cross-attention, feed-forward layer, layer normalization and residual connections is inserted between each sub-layer of the base module.

Different from most previous fine-tuning studies that also adopt the parameter-efficient strategy (\cite{Parameter-efficient-1,Parameter-efficient-2}), the architectures of the adapter modules in ABNet have no need to been fixed. Instead, we can employ different architectures of adapter modules on the encoder and decoder sides, which allows it to be easily adjust to different downstream tasks. As shown in Figure \ref{scneII-abnet-cloud}, the architecture of ABNet is suitable for decoupling. We re-implemented the model in the PaddlePaddle deep learning framework and deployed it on two remote cloud clusters, and named it as ABNet-Cloud. We assume that two pre-trained models from the source and target language domains are distributed in remote cloud cluster $S$ and $T$ separately. Before training, each cluster obtains the corresponding sequences $\mathcal{X}$ and $\mathcal{Y}$ from the parallel training dataset $(\mathcal{X}, \mathcal{Y})$, respectively, which can be controlled and dispatched by the cloud management platform to guarantee users' data privacy. 

During the feed-forward stage, the representations of source sequences are sent from the remote cloud cluster $S$ to the remote cloud cluster $T$ after communication optimization. On the remote cloud cluster $S$, source sequences are fed into the encoder which consists of the pre-trained model from source language domain and inserted adapter module. Formally, we denote the adapter module and the pre-trained model layer block in the encoder side as $\operatorname{AENC}(\cdot)$ and $\operatorname{PLM-S}(\cdot)$, respectively. And the hidden state of each encoder layer in the model is computed as:
\begin{equation}
H_{l+1}^{E}=\operatorname{AENC}\left(\operatorname{PLM-S}\left(H_{l}^{E}\right)\right),
\end{equation}
where $H_{l}^{E}$ denotes the hidden state of the $l$-th encoder layer. After obtaining the hidden state $H^{E}$ of the last encoder layer, we take it as the representations of source sequences, which are sent to the cloud cluster $T$. 

On the remote cloud cluster $T$, target sequences and the representations of source sequences are fed into the decoder which consists of the pre-trained model from the target language domain and inserted adapter module with extra multi-head cross-attention. Specifically, the multi-head cross-attention layers in adapter modules extract conditional context from the representations of source sequences. Formally, we denote the adapter module and the pre-trained model layer block in 
decoder side as $\operatorname{ADEC}(Q,K,V)$ and $\operatorname{PLM-T}(\cdot)$, respectively. And the hidden state of each decoder layer in the model is computed as:
\begin{equation}
H_{l+1}^{D}=\operatorname{ADEC}\left(\operatorname{PLM-T}\left(H_{l}^{D}\right), H^{E}, H^{E}\right),
\end{equation}
where $H_{l}^{D}$ represents the hidden state of the $l$-the decoder layer.

During back propagation, the gradients of the last encoder layer are sent from the remote cloud cluster $T$ to the remote cloud cluster $S$ after communication optimization. On the remote cloud cluster $T$, the decoder computes the conditional MLM loss according to the hypothesis sequences and target sequences, and subsequently returns the gradients of the last encoder layer. On the remote cloud cluster $S$, the encoder continually computes the gradient of each remaining encoder layer. Finally, the inserted adapter modules in en/decoder are trained to adapt for the machine translation task while the parameters of the pre-trained models are frozen during training.

\subsection{Network architecture and parallel computing for both the scenarios}
Some novel parallel computing methods are introduced to train with high performance on different clusters and to compensate for the variance of computational capability of different clusters. 

For Scenario-I, data parallelism was adopted for both the generator side and the discriminator side and we only used one generator for simplification. One problem is that we have to find a proper way to train the generator and the discriminator on different clusters so as to maximize the hardware utility. Since the data flow is only one-way, i.e. the generator $G$ only needs to send the feed-forward result to the discriminator $D$, but the $D$ does not need to send the gradient back to $G$, the basic inter-cluster training architecture for Scenario-I is simple. After $G$ generates a result for a micro-batch, it sends this result to $D$ through the network and does the reset computation. After receiving the result, the discriminator $D$ then carries out the corresponding computation. 

However, this basic idea introduces a challenge. If $G$ just sends the results to the discriminator for the current micro-batch and then moves on to the next micro-batch, $G$ and $D$ might not be in the same training pace, i.e. they may process different micro-batches at the same time. This divergence will get larger when $G$ and $D$ are of different sizes, which might cause the models saved by $G$ and $D$ are different regarding to the amount of training data. To handle this, a synchronization method was added. For Scenario-I, after each optimization stage (updating the parameters in $G$ and $D$), we let $D$ send a small tensor to $G$ to guarantee that these two parts are in the same pace of training.

Besides, the variance of computational capability of different clusters introduces challenge for high performance training. For example, if $G$ is located on a computationally faster cluster and $D$ is located on a slower cluster\footnote{This can be caused by diverse reasons, e.g. the different scales or types of hardware, how a certain deep learning framework behaves on different clusters, etc.}, assigning each generator with one discriminator will result in low hardware utilization on the $G$ side, as $G$ spends more time on syncing with $D$ than on training. To address this challenge, each generator will be assigned with multiple discriminators according to the sizes of the generator and the discriminator. If $G$ is assigned with $n$ $D$s, it will first perform $n$ forward computations once and produce $n$ results. All these $n$ results are sent from $G$ to a root discriminator $D_r$ in the discriminator cluster, then scattered to the corresponding sub-discriminators $D_{subs}$. With this modification, $G$ will spend more time on computing, which can increase the hardware utilization. The synchronization method should be updated thereupon, all the $n$ discriminators $D_{subs}$ should sync first before the $D_r$ syncs with $G$ to make sure each trainer are in the same training pace. In a sentence, $n$ resources on the slower cluster will be connected to $1$ resource on the faster cluster, where the number $n$ is estimated based on the practical performance test on each cluster.

\begin{figure}
  \centering
  \includegraphics[scale=0.43]{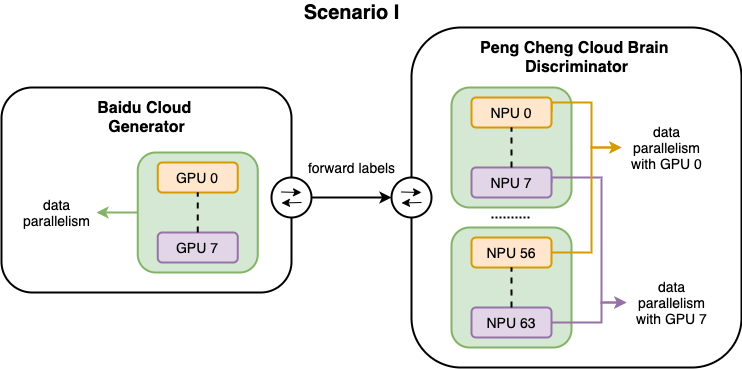}
  \caption{Parallel computing structure for Scenario-I.}
  \label{scneI-paral}
\end{figure}

The overall parallel computing method for Scenario-I is shown in Figure \ref{scneI-paral}. According to our test, the number $n$ was set as $8$, i.e. eight discriminators will be assigned to one generator in our environment. $GPU\ 0$ in the generator side firstly computes eight times in each run and generate eight results. These eight results are then sent to $NPU\ 0$ in the discriminator side. Then $NPU\ 0$ of the discriminator side will scatter the corresponding data to $NPU\ 8$, $NPU\ 16$, $NPU\ 32$, ..., $NPU\ 56$ respectively. After each run, $NPU\ 8$, $NPU\ 16$, $NPU\ 32$, ..., $NPU\ 56$ will sync with $NPU\ 0$ within the discriminator, while $GPU\ 0$ of the generator side will sync with $NPU\ 0$ of the discriminator side. Note that both the generator side and the discriminator side can be scaled to more hardware resources, such as $16$ $GPUs$ with $128$ $NPUs$.

\begin{figure}
  \centering
  \includegraphics[scale=0.43]{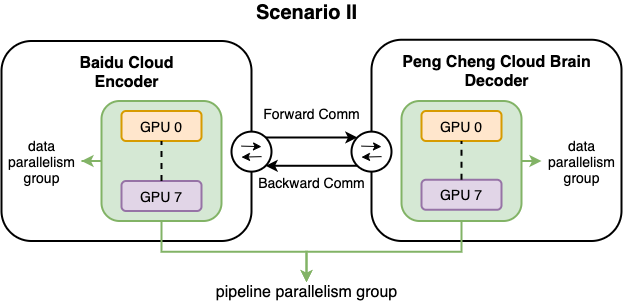}
  \caption{Parallel computing structure for scenario II.}
  \label{scneII-paral}
\end{figure}

For Scenario-II, the encoder not only has to send the results to the decoder, but also has to receive the gradients from the decoder. We adopted the pipeline parallelism provided by the PaddlePaddle framework, which can provide all the send/recv operations needed by Scenario-II. Inside each cluster, data parallelism was adopted. Since the encoder and the decoder are of the same size in Scenario-II, which means that we can assign almost the same scale of hardware for them. The overall parallel computing structure is illustrated in Figure \ref{scneII-paral}.

\section{Experiment}
\subsection{ERNIE-M Extra-Cloud}
% Compare ERNIE-M Extra-Cloud with benchmarks on different multilingual tasks.\\
% Compare ERNIE-M Extra with ERNIE-M Extra-Cloud \\
% Ablation studies for with and without Nebula-CloudOptimizer \\
% Convergence curve for collaborative training \\
% Communication analysis

\paragraph{Pre-training Data} ERNIE-M Extra was trained with monolingual and parallel corpora. For the monolingual data, ERNIE-M Extra adopted the CC100 corpus \cite{cc100,XLM-R} used in XLM-R \cite{XLM-R}, including 116 languages and 5 of which are romanized languages. For the bilingual data, we used a total of 15 languages as INFOXLM \cite{infoxlm}, collected from MultiUN \cite{UN}, IIT Bombay \cite{Bombay}, OPUS \cite{OPUS}, and WikiMatrix \cite{WM}. Following M2M-100 \cite{M2M-100}, we grouped all the training data into 15 groupings by language families.

\paragraph{Pre-training Settings} The prevailing Transformer-encoder was adopted as the backbone of the model in Scenario-I. For the generators, a structure with 12 layers, 768 hidden units, 12 heads was employed. For the discriminator, a structure with 24 layers, 1,024 hidden units, 16 heads was put to use. The activation function used is GeLU \cite{gelu}. In order to  maximize the utility of existed models, we initialized the parameters of the generators with $\textsc{ERNIE-M}_{Base}$, and the discriminator with $\textsc{ERNIE-M}_{Large}$, respectively. We used the Adam optimizer \cite{Adam} to train ERNIE-M Extra; the learning rate was scheduled with a linear decay with 10K warm-up steps, and the peak learning rate was $1e-4$. The hyperparameter $\lambda$ and $\gamma$ were set to 50 and 1 respectively. The training was separated   into two steps, i.e. intra-cluster and inter-cluster. During intra-cluster training, we conducted the pre-training experiments using 64 NVIDIA A100-40GB GPUs with 2,048 batch size and 512 max length. During inter-cluster training, we used 8 NVIDIA V100-32GB GPUs and 64 Ascend 910-32GB NPUs and keep other hyperparameters the same.  

\paragraph{Evaluation Tasks} We executed experiments on the typical cross-lingual evaluation benchmarks XNLI \cite{XNLI} to evaluate the fine-tuning performances of the pre-trained ERNIE-M Extra.
\paragraph{Cross-lingual Natural Language Inference} As a multilingual language inference task, cross-lingual natural language inference (XNLI) task aims to determine the relationship between the two input sentences. It is noteworthy that we only evaluated ERNIE-M Extra in the cross-lingual transfer \cite{XNLI} setting, in which the model is fine-tuned with the English training set and evaluated on the foreign language XNLI test set. 

\renewcommand\arraystretch{0.9}

\begin{table*}[!ht]
\caption{Results on XNLI cross-lingual natural language inference Task. We compare ERNIE-M Extra with 6 previous SoTA baselines including XLM, XLM-R, InfoXLM, Unicoder, VECO and ERNIE-M. The accuracy on each of the 15 XNLI languages and the average accuracy are reported, in which the results of ERNIE-M Extra are based on five runs with different random seeds.}
\centering
\vskip 0.1in
\resizebox{\textwidth}{!}{

\begin{tabular}{l|ccccccccccccccc|c}
\toprule
\textbf{Model}& \textbf{en}& \textbf{fr}& \textbf{es}& \textbf{de}& \textbf{el}& \textbf{bg}& \textbf{ru}& \textbf{tr}& \textbf{ar}& \textbf{vi}& \textbf{th}& \textbf{zh}& \textbf{hi}& \textbf{sw}& \textbf{ur}& \textbf{Avg}  \\
\midrule
\multicolumn{17}{l}{\textit{Fine-tune cross-lingual model on English training set (Cross-lingual Transfer)}} \\
\midrule
XLM \cite{XLM} & 85.0 & 78.7 & 78.9 & 77.8 & 76.6 & 77.4 & 75.3 & 72.5 & 73.1 & 76.1 & 73.2 & 76.5 & 69.6 & 68.4 & 67.3 & 75.1 \\
Unicoder \cite{Unicoder} & 85.1 & 79.0 & 79.4 & 77.8 & 77.2 & 77.2 & 76.3 & 72.8 & 73.5 & 76.4 & 73.6 & 76.2 & 69.4 & 69.7 & 66.7 & 75.4 \\
XLM-R \cite{XLM-R} & 85.8 & 79.7 & 80.7 & 78.7 & 77.5 & 79.6 & 78.1 & 74.2 & 73.8 & 76.5 & 74.6 & 76.7 & 72.4 & 66.5 & 68.3 & 76.2 \\
\textsc{InfoXLM} \cite{infoxlm} & 86.4 & 80.6 & 80.8 & 78.9 & 77.8 & 78.9 & 77.6 & 75.6 & 74.0 & 77.0 & 73.7 & 76.7 & 72.0 & 66.4 & 67.1 & 76.2 \\
\textsc{Ernie-M} \cite{ERNIE-M} & 85.5 & 80.1 & 81.2 & 79.2 & 79.1 & 80.4 & 78.1 & 76.8 & 76.3 & 78.3 & 75.8 & 77.4 & 72.9 & 69.5 & 68.8 & 77.3 \\

\midrule
XLM-R$_{\scriptsize \textsc{Large}}$ \cite{XLM-R} & 89.1 & 84.1 & 85.1 & 83.9 & 82.9 & 84.0 & 81.2 & 79.6 & 79.8 & 80.8 & 78.1 & 80.2 & 76.9 & 73.9 & 73.8 & 80.9 \\
$\textsc{InfoXLM}_{\scriptsize \textsc{Large}}$ \cite{infoxlm} & \textbf{89.7} & 84.5 & 85.5 & 84.1 & 83.4 & 84.2 & 81.3 & 80.9 & 80.4 & 80.8 & 78.9 & 80.9 & 77.9 & 74.8 & 73.7 & 81.4 \\
VECO$_{\scriptsize \textsc{Large}}$ \cite{VECO} & 88.2 & 79.2 & 83.1 & 82.9 & 81.2 & 84.2 & \textbf{82.8} & 76.2 & 80.3 & 74.3 & 77.0 & 78.4 & 71.3 & \textbf{80.4} & \textbf{79.1} & 79.9 \\
$\textsc{Ernie-M}_{\scriptsize \textsc{Large}}$ \cite{ERNIE-M} & 89.3 & \textbf{85.1} & 85.7 & 84.4 & 83.7 & 84.5 & 82.0 & 81.2 & 81.2 & \textbf{81.9} & 79.2 & 81.0 & 78.6 & 76.2 & 75.4 & 82.0 \\
ERNIE-M Extra & 89.4 & \textbf{85.1} & \textbf{86.0} & \textbf{84.5} & \textbf{84.4} & \textbf{84.6} & 81.8 & \textbf{81.7} & \textbf{81.8} & \textbf{81.9} & \textbf{79.3} & \textbf{81.2} & \textbf{79.1} & 76.3 & 75.7 & \textbf{82.2} \\
\bottomrule

\end{tabular}}
\label{xnli}
\vskip -0.1in
\end{table*}

\paragraph{Results} The results of ERNIE-M Extra on the XNLI task are reported in Table \ref{xnli}. As shown in Table \ref{xnli}, ERNIE-M Extra outperforms all baseline models including XLM \cite{XLM}, Unicoder \cite{Unicoder}, XLM-R \cite{XLM-R}, \textsc{InfoXLM} \cite{infoxlm}, VECO \cite{VECO} and ERNIE-M \cite{ERNIE-M} on most of languages. The reported scores on the test set are averaged over five runs with different random seeds. Under cross-lingual transfer setting, ERNIE-M Extra achieves 82.2 accuracy, outperforming \textsc{InfoXLM}$_{\scriptsize \textsc{Large}}$ by 0.8 and slightly better than  $\textsc{Ernie-M}_{\scriptsize \textsc{Large}}$.

We trained ERNIE-M Extra on both the intra-cloud and inter-cloud environment. The network optimizations on multi clouds are illustrated in \S \ref{network}.

\subsection{ABNet-Cloud}
\label{section:Exper-ABNet-Cloud}

\paragraph{Datasets} 
For Scenario-II, we conducted experiments on Español-English (Es-En) and Chinese-English (Zh-En) translations from IWSLT'14 datasets \footnote{https://wit3.fbk.eu/} to verify the effectiveness of the machine translation model for our fine-tuning task, which has been replicated by us using PaddlePaddle from PyTorch, and with the multilingual model replaced by ERNIE-M \cite{ERNIE-M}. For all language pairs in IWSLT'14, we merged the validation dataset dev 2010 and the test datasets tst 2010, tst 2011, tst 2012. And we reported the BLEU score on the merged dataset, which is the same as the dataset configurations of ABNet strictly. Preprocessing like tokenization was done automatically with the sentencepiece or wordpiece program, which depends on the pre-trained model.
\paragraph{Metric}
Following the metric configuration of ABNet, we used case-insensitive BLEU as the evaluation metric. The BLEU score is calculated using the \texttt{multi-bleu.perl}.
\paragraph{Implementations}
We used the ERNIE-M-base model \cite{ERNIE-M}, mBERT-base model and BERT-base model \cite{BERT} as the pre-trained language models in our experiments. Specifically, for the encoder side we use ERNIE-M-base-cased or mBERT-base-cased as the pre-trained model from the source language domain. For the decoder side, We used bert-base-uncased as the pre-trained model from the target language domain. For adapter modules, we adopt 512 dimensions for the hidden state between two feed forward layers on the encoder side. On the decoder side, we adopt 768 dimensions for the hidden state of the cross attention module, which is equal to the hidden dimension of BERT-base models. We trained the model with a batch size of 128 sequences and 64 max length. Parameters were optimized by using Adam optimizer \cite{Adam}, with $\beta_{1}=0.9$, $\beta_{1}=0.98$ and $\epsilon = 1 \times 10^{-8}$, with $\text{warmup\_steps}=4000$. Label smoothing \cite{Label-smoothing} of value $=1$ is also adopted. The training was separated into two steps, i.e. intra-cluster and inter-cluster. During intra-cluster training, we conducted the fine-tuning experiments using 2 NVIDIA V100-32GB GPUs. During inter-cluster training, we deployed the encoder and decoder on 8 NVIDIA V100-32GB GPUs for each cluster. For the inference, we used the beam search algorithm with $\text{beam\_sizes}=4$ to obtain the translation from the ABNet model.
\begin{table}[!ht]
\begin{center}
\caption{The BLEU scores of ABNet and the Transformer-base model on IWSLT'14 Es-En and Zh-En tasks. "-" indicates the corresponding experiment was not conducted.}
\begin{tabular}{cccccc}
   \toprule
   \textbf{Method} & \textbf{Es-En} & \textbf{Zh-En} \\
   \midrule
   Transformer-base \cite{Transformer} & 39.60 & 13.17 \\
   $\textup{ABNet}_{\scriptsize \textup{mBERT-BERT}}$ & 43.01 & 14.12  \\
   $\textup{ABNet}_{\scriptsize \textup{ERNIE-M-BERT}}$ (ours) & 43.19 & 17.87  \\
   \midrule
   $\textup{ABNet-Cloud}_{\scriptsize \textup{ERNIE-M-BERT}}$ (ours) & 43.77 & - \\
   \bottomrule
   \label{scneII-experiment}
\end{tabular}
\end{center}
\end{table}
\subparagraph{Results} Table \ref{scneII-experiment} shows the results for the Es-En and Zh-En translation tasks. We compared our re-implemented ABNet with the Transformer-base model in the PaddlePaddle deep learning framework. As can be observed from Table \ref{scneII-experiment}, with employing pre-trained models on both the encoder and decoder sides, our methods obtain significant improvements on both language pairs compared with Transformer-base. This validates the correctness of our re-implementation and the effectiveness of ABNet to utilize the knowledge from pre-trained language models for better translation. We observed that the performance of $\textup{ABNet}_{\scriptsize \textup{ERNIE-M-BERT}}$ is better than $\textup{ABNet}_{\scriptsize \textup{mBERT-BERT}}$ on both language pairs, especially on the Zh-En pair. This phenomenon can be mainly attributed to the superiority of ERNIE-M compared with mBERT. An interesting observation is that the performance of $\textup{ABNet-Cloud}_{\scriptsize \textup{ERNIE-M-BERT}}$ is slightly better than $\textup{ABNet}_{\scriptsize \textup{ERNIE-M-BERT}}$. This signifies that the appropriate communication optimization might even bring a slight increase in accuracy while significantly improve the efficiency of transmission.

% Different from most previous fine-tuning studies that also adopt the parameter-efficient strategy, the architectures of the adapters in ABNet have no need to been fixed. Instead, we can employ different architectures of adapters on the encoder and decoder sides, which allows it to be easily adjust to different downstream tasks. Additionally, we re-implemented ABNet in PaddlePaddle deep learning framework and deployed it on two cloud clusters, and named it as ABNet-Cloud. Some contrast experiments are conducted regarding the performance of ABNet-Cloud in Section \ref{section:Exper-ABNet-Cloud}. We can observed that ABNet-Cloud with communication optimization significantly outperforms the Transformer-base model.

% BLEU on two or three machine translation tasks of ABNet-Cloud, compared with Transformer-base, including ES-EN, EN-ES + zh-and two low resource languages \\
% Ablation studies ABNet-Cloud with and without Nebula-CloudOptimizer \\
% Convergence curve for collaborative training \\
% Communication analysis

\subsection{Network, parallel computing and communication experiment}
\label{network}
To test the performance of the framework, we conducted some experiments on the throughput of Scenario-I.  To test the intra-cloud training, we carried out the test on Peng Cheng Cloud Brain I (V100). To test the inter-cloud training with the homogeneous hardware structure, we carried out the test on Baidu Cloud (V100) and Peng Cheng Cloud Brain I (V100). To test the inter-cloud training with heterogeneous hardware structure, we carried out the test on Baidu Cloud (V100) and Peng Cheng Cloud Brain II (Ascend 910). The experiments results for homogeneous hardware can be found in Table \ref{trhoughput_same_structure}, and the results for heterogeneous hardware can be found in Table \ref{trhoughput_diff_structure}.

\begin{table}[ht]
\begin{center}
\caption{Throughput for Homogeneous Hardware}
\label{trhoughput_same_structure}
\begin{tabular}{ccc}
\toprule
 Cloud Cluster(s) & Throughput Ratio  \\
 \midrule
 Intra Peng Cheng Cloud Brain I &   1.0   \\
 Inter Baidu Cloud and Peng Cheng Cloud Brain I  &  0.85 \\
 \bottomrule
\end{tabular}
\end{center}
\end{table}
\begin{table}[ht]
\begin{center}
\caption{Throughput for Heterogeneous Hardware}
\label{trhoughput_diff_structure}
\begin{tabular}{ccc}
\toprule
 Cloud Clusters & Speedup Ratio  \\
 \midrule
 Inter Baidu Cloud and Peng Cheng Cloud Brain II (8 NPUs) & 1.0 \\
 Inter Baidu Cloud and Peng Cheng Cloud Brain II (64 NPUs) & 4.34 \tablefootnote{Some optimizations could be further applied for improving the performance.} \\
 \bottomrule
\end{tabular}
\small
\end{center}
\end{table}

\begin{figure}
  \centering
  \includegraphics[scale=0.6]{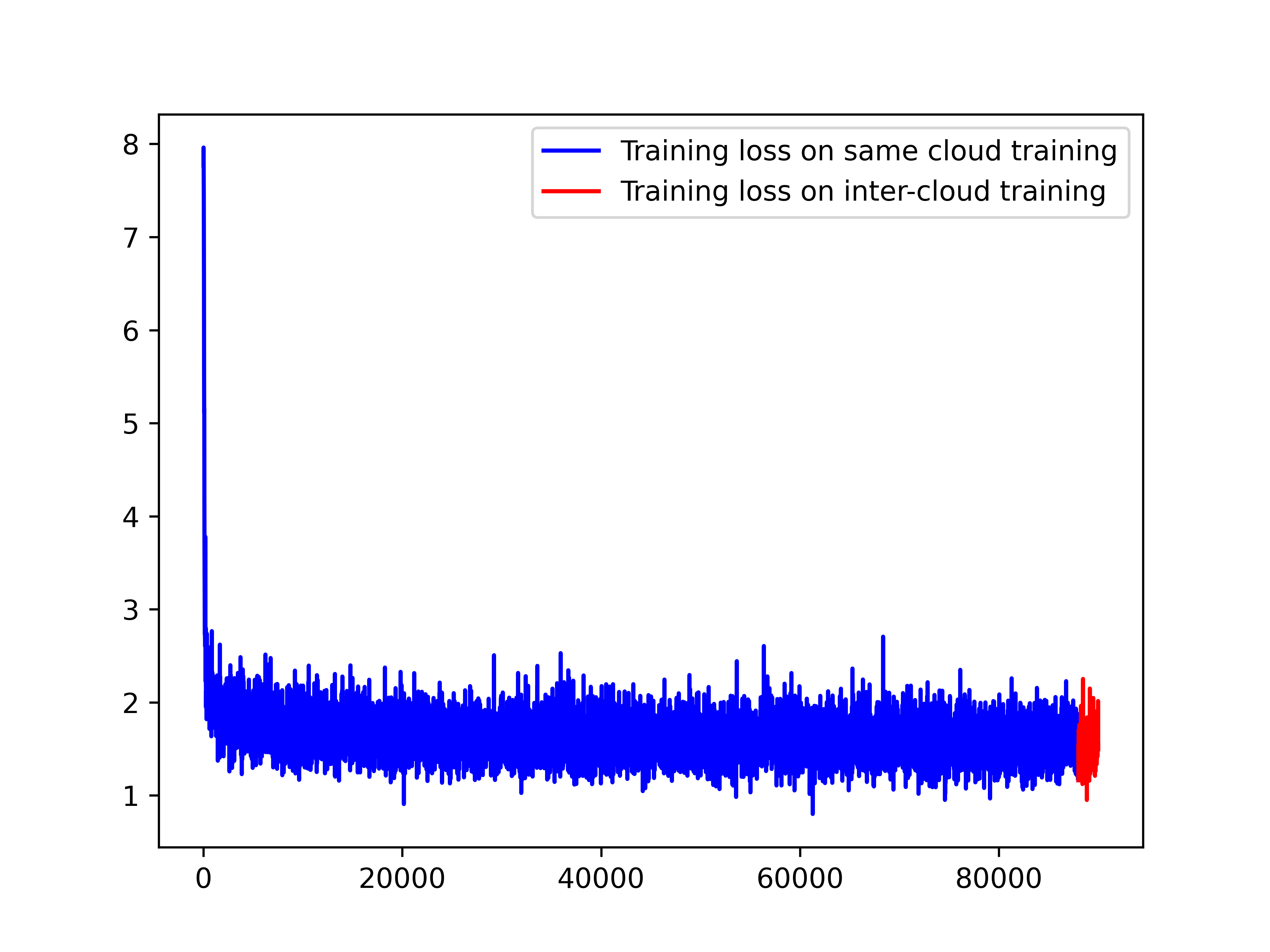}
  \caption{Training loss for scenario I.}
  \label{scneI-loss}
\end{figure}

Beside the performance experiments, we also carried out accuracy experiment for inter-cloud training. Considering the fact that the inter-cloud resources are limited, for the inter-cloud training, we hot start the model with a checkpoint instead of training model from scratch. In the experiment, we first trained the model for Scenario-I on intra-cloud for 88,000 steps and then resume the training on the inter-cloud environment (Baidu Cloud and Peng Cheng Cloud Brain I) for another 2,000 steps. From the training loss shown in Figure \ref{scneI-loss}, we can see that the inter-cloud training won't effect the convergence of the model.

\begin{figure}[H]
  \centering
  \includegraphics[scale=0.45]{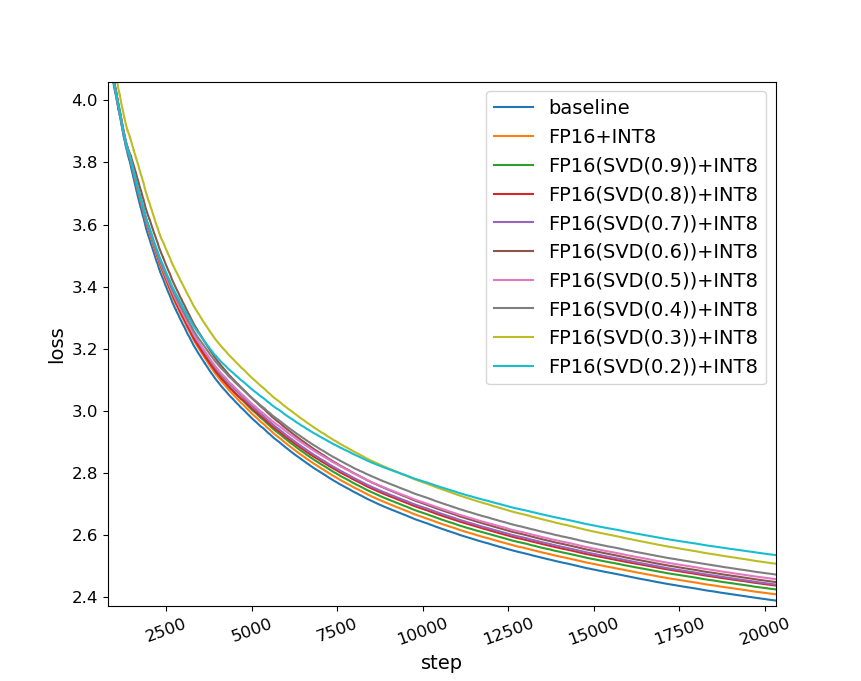}
  \caption{Training loss with different compression methods.}
  \label{lossfig}
\end{figure}

To test the effect of communication optimization, we conducted experiments using different communication compression  strategies in Scenario II, i.e. the $\textup{ABNet-Cloud}_{\scriptsize \textup{ERNIE-M-BERT}}$ Es-En fine-tuning task. The experiment setting was the same as that in \S \ref{section:Exper-ABNet-Cloud}. FP16 quantization was used for feed-forward activation compression before transmission, and INT8 quantization was used for back propagation gradient compression before transmission. SVD compression was nested with FP16 to further reduce the communication volume following, and the compression ratio effect was tested by choosing different SVD singular values ratios. The experiments were carried out on Baidu Cloud (V100) and Peng Cheng Cloud Brain I (V100), which was connected by Internet with bandwidth up to 60 Mbit/s. The test results of models training with different compression communication methods are shown in Table \ref{CompressBleu}, and the training loss of each method is shown in Figure \ref{lossfig}.

\begin{table}[ht]\label{table:compressionratiovsspeed}
\begin{center}
\caption{Effects of different compression methods for Scenario-II on Es-En machine translation task}
\label{CompressBleu} 
\begin{tabular}{ccccc}
\toprule
 \textbf{Compression method} & \textbf{BLEU} & \makecell{\textbf{Forward} \\ \textbf{compression ratio}} & \makecell{\textbf{Backward} \\ \textbf{compression ratio}} & \makecell{\textbf{Training speed} \\ \textbf{(s/step)}}  \\
 \midrule
baseline&43.19 &1.00 &1.00 &4.42 \\
FP16+INT8&38.82 &0.50 &0.25 &1.60 \\
FP16(SVD(0.9))+INT8&39.71 &0.45 &0.25 &2.10 \\
FP16(SVD(0.8))+INT8&41.15 &0.40 &0.25 &1.50 \\
FP16(SVD(0.7))+INT8&40.74 &0.34 &0.25 &1.56 \\
FP16(SVD(0.6))+INT8&41.92 &0.30 &0.25 &1.42 \\
FP16(SVD(0.5))+INT8&39.45 &0.25 &0.25 &1.24 \\
FP16(SVD(0.4))+INT8&39.56 &0.20 &0.25 &0.94 \\
FP16(SVD(0.3))+INT8&37.48 &0.15 &0.25 &0.90 \\
FP16(SVD(0.2))+INT8&39.00 &0.09 &0.25 &0.86 \\
 \bottomrule
\end{tabular}
\end{center}
\end{table}

In Table \ref{CompressBleu}, the baseline was trained with no-compression communication, FP16+INT8 was trained with feed-forward FP16 compression and back propagation INT8 compression, FP16(SVD($r$))+INT8 was trained with feed-forward FP16 and SVD($r$) nested compression and back propagation INT8 compression, and $r$ is the used ratio of the total singular values. We compared the effect of different compression ratios on the final BLEU results and training speed. The communication compression was used from the first training step in Table \ref{CompressBleu}.

From the experimental results in Table \ref{CompressBleu}, it can be seen that the BLEU values of all the methods with communication compression are lower than the baseline, among which the best BLEU is 41.92. This demonstrates that using compression communication will lead to loss of model accuracy. The comparison of different compression methods, however, shows that more data compressed did not always lead to more BLEU loss. For example, comparing with FP16+INT8, FP16(SVD(0.2))+INT8 compresses 41\% more data but improves 0.18 in BLEU. FP16(SVD(0.6))+INT8 with the compression ratio 0.30 is the best BLEU in all the compression methods. This signifies that the compression ratio can be taken as a hyper-parameter for tuning for communication optimization. 

From Table \ref{CompressBleu} we can see that, data compression can reduce the data transmission time and speed up the model training. FP16(SVD(0.2))+INT8 reduces 3.56 seconds for each training step compared with the baseline. It only sends 17\% data of the baseline and uses 19\% time. The results prove that communication compression is effective to reduce the training time in the  cloud-cluster environment. FP16(SVD(0.9))+INT8 sends less data but the speed is 0.5 seconds slower than FP16+INT8. It shows that the SVD decomposition is likely to consume much time on computing. If the data are not compressed enough, the training speed will be slower. Thus, the selection of the compression ratio is an important step.

From the training losses in Figure \ref{lossfig}, we can see that using communication compression will lead to training loss increase, and after about 8,000 steps, more compression leads to larger training loss. All the compression methods, however, still can guarantee the convergence of model training.

\begin{figure}[H]
  \centering
  \includegraphics[scale=0.25]{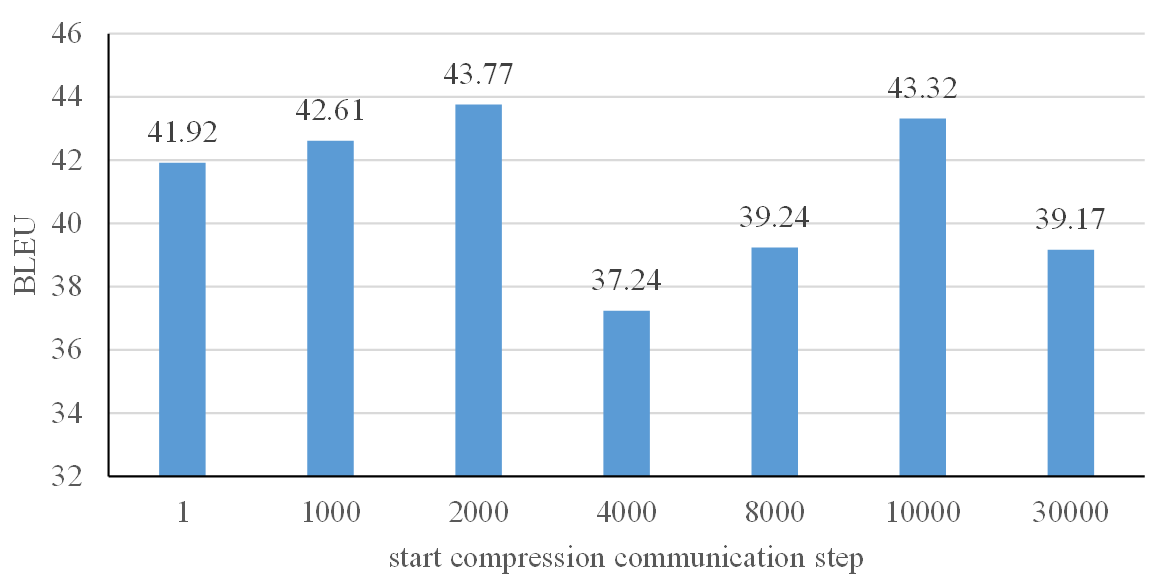}
  \caption{Performance of using FP16(SVD(0.6))+INT8 from different steps.}
  \label{different_start_step}
\end{figure}

We also conducted experiments on starting communication compression from different training steps. We chose FP16(SVD(0.6))+INT8, the communication compression method with the best performance according to Table \ref{CompressBleu}, as the targeted method. The experimental results are shown in Figure \ref{different_start_step}. We can observe that starting using communication compression from step 1,000, 2,000 and 10,000 can further improve the BLEUs, and from step 2,000 and step 10,000, the performance are even better than the baseline in Table \ref{CompressBleu}. This means that if choosing an appropriate startup step to use compression, it can further balance the accuracy and communication efficiency.

\section{Discussion}
\paragraph{Overview}
The experimental results demonstrate that the NLP tasks in the two scenarios are practical running between the two clusters. For Scenario-I, we have shown that when the communication traffic is low, the inter-cloud training did not affect much of the throughput, which can be accepted. Furthermore, when hot starting the model in the intra-cluster environment and then resuming the training over clouds, we found that the loss can still converge well. For Scenario-II, since the communicated volume is much larger than that of Scenario-I, we have applied several optimization techniques to enhance the parallization degree and accelerate the communication, results demonstrate that with the help of Nebula-I, the training speed can be considerably improved, i.e. around 3$\times$ compared with the baseline while preserving satsifactory performance (Table \ref{CompressBleu}). It is interesting to see from Figure \ref{different_start_step} that after performing compression operations even can improve the BLEU of the baseline (43.77 vs 43.19). We deem that the compression by giving up some data is similar to the Dropout mechanism \cite{Dropout} that is widely applied in deep learning, which can somehow improve the generalizability of a model. 
% Discuss the experiment results, including how ELECTRA and ABNet behave, performance of multiple layers.

\paragraph{Trade-off between accuracy and efficiency}
The training environment cross-cloud clusters encounters the low-bandwidth and high-latency problem, which would significantly slow down the training process. Though we have multiple compression techniques to reduce the communication time during training, it may sacrifice some accuracy when the compression ratio is not properly used. Furthermore, according to our experimental results (Table \ref{CompressBleu}), some high compression ratios may achieve better generalization performance (possibly similar to the dropout mechanism). Thus, how to select a proper compression ratio (or a compression method) for a particular training task to achieve the best performance under specific cross-cloud environments is worth for further exploration. 
% Cloud-based learning tasks may suffers from communication problems such as high latency, intermittent connections and low network throughput \cite{eSGD, Deep-gradient-compression}. These problems are converging with some of those mentioned in previous research areas such as federated learning \cite{FL-outlook} and edge computing \cite{Gaia}. However, as our compute nodes are clusters that have much larger computational capacities rather than edge computers or mobile devices, we are allowed to exploit tasks that require larger scales of compute such as pre-training and fine-tuning associated with big data and large models.

% Previous studies have demonstrated that most communication compression techniques would lead to some accuracy loss \cite{Compression-survey}. However, there should be a balance between efficient training and sacrificing the accuracy. For example, when preferring the accuracy boosting to the training efficiency, i.e. users can wait a bit but require an improved accuracy, we can tune down the compression rate of the communication load. Otherwise, we have to find effective ways that maximizing the training speed while minimizing the accuracy loss. These selections might be especially necessary when facing the cloud-based training scenarios where the communication is the primary bottleneck.

\paragraph{Hardware heterogeneity}
Unlike clusters located on a data center, cross-cloud clusters generally have different hardware configurations (i.e., hardware heterogeneity). The hardware heterogeneity easily makes the training speed unbalanced as some slow processors would become stragglers resulting in inefficient training. For example, in our studied environment, one cloud is equipped with GPUs and the other is with NPUs. Though the straggler problem in distributed training has been well studied, our proposed solutions (i.e., splitting the model to two parts according to the model architecture) to enable distributed training in the cross-cloud environment is quite different from the data-parallel scenario~\cite{straggler}. How to split the model and displace the model to different cloud clusters by considering the hardware configurations to optimally utilize the hardware resources is another direction for further study ~\cite{End-to-End}. 

% The parallelism strategy needs to be redesigned according to different hardware, e.g. from GPU to NPU. The PaddlePalle framework supports the adaptive re-partition between heterogeneous hardware to maximize the utility of computation architectures. For example, when transferring a deep learning model from GPU to NPU, PaddlePaddle can split task-specific layers to automatically adapt to the network topology and software stack of the NPU cluster \cite{ERNIE-Titan}. With this functionality, the program can easily run on heterogeneous platforms. However, we still face the throughput problem where the NPU's throughput, e.g. processed tokens/s, is weaker than GPU's, when running PaddlePaddle programs. Thus, we need to take the throughput into consideration when designing the optimization strategies. 
\paragraph{Security}
In the current version of Nebula-I, we only provide basic security mechanisms for computation and communication based on the trust between two cross-cloud clusters. However, when deploying the system on two clouds that have not any trustworthy mechanisms, the communicated data (e.g., activation outputs) cross two clouds may leak the data privacy. It should also be carefully designed when the data cannot be shared across the cloud clusters. 

In addition to the Nebula-I framework that proposed to optimize the cloud-based training, we also would like to promote more on the utility of pre-trained models by means of showing the effectiveness of the models in our two scenarios. Currently, the re-use of many large models still have limitations. For example, some of the models only provide inference APIs while most end users do not have the hardware capacity to hold them \cite{GPT-3, PANGU, ERNIE-Titan}. Scenario-I is designed to re-use existed models to assist the pre-training of a model with larger capabilities. Figure \ref{scneI-ernie-m-cloud} shows a method to aggregate the ability of different pre-trained models, which is theoretically effective and efficient. This can also be scaled out to other tasks where each language cluster can be replaced by a teacher. Each teacher in different clusters has its own expertise and can teach the student collaboratively. In this case, more machine learning tasks can be integrated. Scenario-II is also a good case in which pre-trained models provide their knowledge to downstream tasks. We believe that by deploying large models onto cloud clusters and adding task-specific accessories (e.g. lightweight parameters), the values of them can be further amplified.

\paragraph{Conclusion}
In this work, we proposed a general framework, Nebula-I, which is implemented using the PaddlePaddle deep learning framework, for collaboratively training deep learning models over remote heterogeneous clusters. We applied Nebula-I in two different NLP scenarios which include pre-training a multilingual language model and fine-tuning a machine translation model. Results demonstrate that both model accuracy and communication efficiency can be satisfied with the help of Nebula-I. The success in these scenarios not only shows the effectiveness of Nebula-I in optimizing the whole deep learning training process, but also offers potential ways for the reuse and promotion of large models, which would be helpful to the research field. We hope that users can quickly deploy training tasks over cloud clusters under the framework of Nebula-I with minimum development, e.g. by adding several lines of codes and modifying a configuration file. However, training a general model on cloud clusters remains a huge challenge and needs more explorations from multiple perspectives.

\begin{ack}
% Use unnumbered first level headings for the acknowledgments. All acknowledgments
% More information about this disclosure can be found at: \url{https://neurips.cc/Conferences/2022/PaperInformation/FundingDisclosure}.
We sincerely thank Dr. Shujian Huang and Dr. Xiaoxiong Zhong for their valuable suggestions in this work, and thank all the members from Peng Cheng-Baidu NLP Joint Lab for their continuous support.
\end{ack}

\bibliographystyle{nips}
\bibliography{mybib.bib}

\begin{thebibliography}{10}

\bibitem{BERT}
Devlin, J., M.-W. Chang, K.~Lee, et~al.
\newblock {BERT}: Pre-training of deep bidirectional transformers for language
  understanding.
\newblock In \emph{Proceedings of the 2019 Conference of the North {A}merican
  Chapter of the Association for Computational Linguistics: Human Language
  Technologies}. 2019.

\bibitem{GPT-2}
Radford, A., J.~Wu, R.~Child, et~al.
\newblock Language models are unsupervised multitask learners.
\newblock \emph{OpenAI blog}, 2019.

\bibitem{GPT-3}
Brown, T., B.~Mann, N.~Ryder, et~al.
\newblock Language models are few-shot learners.
\newblock \emph{Advances in neural information processing systems}, 2020.

\bibitem{CPM-2}
Zhang, Z., Y.~Gu, X.~Han, et~al.
\newblock Cpm-2: Large-scale cost-effective pre-trained language models.
\newblock \emph{AI Open}, 2021.

\bibitem{PANGU}
Zeng, W., X.~Ren, T.~Su, et~al.
\newblock Pangu-$\alpha$: Large-scale autoregressive pretrained chinese
  language models with auto-parallel computation.
\newblock \emph{arXiv preprint arXiv:2104.12369}, 2021.

\bibitem{ERNIE-Titan}
Wang, S., Y.~Sun, Y.~Xiang, et~al.
\newblock Ernie 3.0 titan: Exploring larger-scale knowledge enhanced
  pre-training for language understanding and generation.
\newblock \emph{arXiv preprint arXiv:2112.12731}, 2021.

\bibitem{Gopher}
Rae, J.~W., S.~Borgeaud, T.~Cai, et~al.
\newblock Scaling language models: Methods, analysis \& insights from training
  gopher.
\newblock \emph{arXiv preprint arXiv:2112.11446}, 2021.

\bibitem{Switch}
Fedus, W., B.~Zoph, N.~Shazeer.
\newblock Switch transformers: Scaling to trillion parameter models with simple
  and efficient sparsity.
\newblock \emph{arXiv preprint arXiv:2101.03961}, 2021.

\bibitem{M6}
Lin, J., R.~Men, A.~Yang, et~al.
\newblock M6: A chinese multimodal pretrainer.
\newblock \emph{arXiv preprint arXiv:2103.00823}, 2021.

\bibitem{PaLM}
Chowdhery, A., S.~Narang, J.~Devlin, et~al.
\newblock Palm: Scaling language modeling with pathways.
\newblock \emph{arXiv preprint arXiv:2204.02311}, 2022.

\bibitem{PS}
Wei, J., W.~Dai, A.~Qiao, et~al.
\newblock Managed communication and consistency for fast data-parallel
  iterative analytics.
\newblock In \emph{Proceedings of the Sixth ACM Symposium on Cloud Computing}.
  2015.

\bibitem{Poseidon}
Zhang, H., Z.~Zheng, S.~Xu, et~al.
\newblock Poseidon: An efficient communication architecture for distributed
  deep learning on $\{$GPU$\}$ clusters.
\newblock In \emph{2017 USENIX Annual Technical Conference (USENIX ATC 17)}.
  2017.

\bibitem{Mesh}
Shazeer, N., Y.~Cheng, N.~Parmar, et~al.
\newblock Mesh-tensorflow: Deep learning for supercomputers.
\newblock \emph{Advances in neural information processing systems}, 2018.

\bibitem{Megatron}
Shoeybi, M., M.~Patwary, R.~Puri, et~al.
\newblock Megatron-lm: Training multi-billion parameter language models using
  model parallelism.
\newblock \emph{arXiv preprint arXiv:1909.08053}, 2019.

\bibitem{Pipedream}
Harlap, A., D.~Narayanan, A.~Phanishayee, et~al.
\newblock Pipedream: Fast and efficient pipeline parallel dnn training.
\newblock \emph{arXiv preprint arXiv:1806.03377}, 2018.

\bibitem{Memory-efficient-pp}
Narayanan, D., A.~Phanishayee, K.~Shi, et~al.
\newblock Memory-efficient pipeline-parallel dnn training.
\newblock In \emph{International Conference on Machine Learning}. PMLR, 2021.

\bibitem{DeepSpeed}
Rasley, J., S.~Rajbhandari, O.~Ruwase, et~al.
\newblock Deepspeed: System optimizations enable training deep learning models
  with over 100 billion parameters.
\newblock In \emph{Proceedings of the 26th ACM SIGKDD International Conference
  on Knowledge Discovery \& Data Mining}. 2020.

\bibitem{Public-Cloud}
Shi, S., X.~Zhou, S.~Song, et~al.
\newblock Towards scalable distributed training of deep learning on public
  cloud clusters.
\newblock \emph{Proceedings of Machine Learning and Systems}, 2021.

\bibitem{Transformer}
Vaswani, A., N.~Shazeer, N.~Parmar, et~al.
\newblock Attention is all you need.
\newblock \emph{Advances in neural information processing systems}, 2017.

\bibitem{topk}
Aji, A.~F., K.~Heafield.
\newblock Sparse communication for distributed gradient descent.
\newblock \emph{arXiv preprint arXiv:1704.05021}, 2017.

\bibitem{qsgd}
Alistarh, D., D.~Grubic, J.~Li, et~al.
\newblock Qsgd: Communication-efficient sgd via gradient quantization and
  encoding.
\newblock \emph{Advances in Neural Information Processing Systems}, 2017.

\bibitem{Compression-survey}
Xu, H., C.-Y. Ho, A.~M. Abdelmoniem, et~al.
\newblock Compressed communication for distributed deep learning: Survey and
  quantitative evaluation, 2020.

\bibitem{FL-outlook}
Ding, J., E.~Tramel, A.~K. Sahu, et~al.
\newblock Federated learning challenges and opportunities: An outlook.
\newblock In \emph{ICASSP 2022-2022 IEEE International Conference on Acoustics,
  Speech and Signal Processing (ICASSP)}. IEEE, 2022.

\bibitem{ELECTRA}
Clark, K., M.-T. Luong, Q.~V. Le, et~al.
\newblock Electra: Pre-training text encoders as discriminators rather than
  generators.
\newblock \emph{arXiv preprint arXiv:2003.10555}, 2020.

\bibitem{KD-MT}
Weng, R., H.~Yu, S.~Huang, et~al.
\newblock Acquiring knowledge from pre-trained model to neural machine
  translation.
\newblock In \emph{Proceedings of the AAAI Conference on Artificial
  Intelligence}. 2020.

\bibitem{ABNet}
Guo, J., Z.~Zhang, L.~Xu, et~al.
\newblock Incorporating bert into parallel sequence decoding with adapters.
\newblock \emph{Advances in Neural Information Processing Systems}, 2020.

\bibitem{Prompt-MT}
Tan, Z., X.~Zhang, S.~Wang, et~al.
\newblock Msp: Multi-stage prompting for making pre-trained language models
  better translators.
\newblock \emph{arXiv preprint arXiv:2110.06609}, 2021.

\bibitem{Geeps}
Cui, H., H.~Zhang, G.~R. Ganger, et~al.
\newblock Geeps: Scalable deep learning on distributed gpus with a
  gpu-specialized parameter server.
\newblock In \emph{Proceedings of the eleventh european conference on computer
  systems}. 2016.

\bibitem{Pathways}
Barham, P., A.~Chowdhery, J.~Dean, et~al.
\newblock Pathways: Asynchronous distributed dataflow for ml.
\newblock \emph{arXiv preprint arXiv:2203.12533}, 2022.

\bibitem{atomo}
Wang, H., S.~Sievert, S.~Liu, et~al.
\newblock Atomo: Communication-efficient learning via atomic sparsification.
\newblock \emph{Advances in Neural Information Processing Systems}, 31, 2018.

\bibitem{powersgd}
Vogels, T., S.~P. Karimireddy, M.~Jaggi.
\newblock Powersgd: Practical low-rank gradient compression for distributed
  optimization.
\newblock \emph{Advances in Neural Information Processing Systems}, 2019.

\bibitem{8bit}
Dettmers, T.
\newblock 8-bit approximations for parallelism in deep learning.
\newblock \emph{arXiv preprint arXiv:1511.04561}, 2015.

\bibitem{lowranksurvey}
Kishore~Kumar, N., J.~Schneider.
\newblock Literature survey on low rank approximation of matrices.
\newblock \emph{Linear and Multilinear Algebra}, 2017.

\bibitem{fl-security1}
Bhowmick, A., J.~Duchi, J.~Freudiger, et~al.
\newblock Protection against reconstruction and its applications in private
  federated learning.
\newblock \emph{arXiv preprint arXiv:1812.00984}, 2018.

\bibitem{fl-security2}
Phong, L.~T., Y.~Aono, T.~Hayashi, et~al.
\newblock Privacy-preserving deep learning via additively homomorphic
  encryption.
\newblock \emph{IEEE Transactions on Information Forensics and Security}, 2018.

\bibitem{ERNIE-M}
Ouyang, X., S.~Wang, C.~Pang, et~al.
\newblock Ernie-m: enhanced multilingual representation by aligning
  cross-lingual semantics with monolingual corpora.
\newblock \emph{arXiv preprint arXiv:2012.15674}, 2020.

\bibitem{XLM-E}
Chi, Z., S.~Huang, L.~Dong, et~al.
\newblock Xlm-e: cross-lingual language model pre-training via electra.
\newblock \emph{arXiv preprint arXiv:2106.16138}, 2021.

\bibitem{M2M-100}
Fan, A., S.~Bhosale, H.~Schwenk, et~al.
\newblock Beyond english-centric multilingual machine translation.
\newblock \emph{Journal of Machine Learning Research}, 2021.

\bibitem{COCO-LM}
Meng, Y., C.~Xiong, P.~Bajaj, et~al.
\newblock Coco-lm: Correcting and contrasting text sequences for language model
  pretraining.
\newblock \emph{Advances in Neural Information Processing Systems}, 2021.

\bibitem{Parameter-efficient-1}
Houlsby, N., A.~Giurgiu, S.~Jastrzebski, et~al.
\newblock Parameter-efficient transfer learning for nlp.
\newblock In \emph{International Conference on Machine Learning}, pages
  2790--2799. PMLR, 2019.

\bibitem{Parameter-efficient-2}
Wang, R., D.~Tang, N.~Duan, et~al.
\newblock K-adapter: Infusing knowledge into pre-trained models with adapters.
\newblock \emph{arXiv preprint arXiv:2002.01808}, 2020.

\bibitem{cc100}
Wenzek, G., M.-A. Lachaux, A.~Conneau, et~al.
\newblock Ccnet: Extracting high quality monolingual datasets from web crawl
  data.
\newblock \emph{arXiv preprint arXiv:1911.00359}, 2019.

\bibitem{XLM-R}
Conneau, A., K.~Khandelwal, N.~Goyal, et~al.
\newblock Unsupervised cross-lingual representation learning at scale.
\newblock \emph{arXiv preprint arXiv:1911.02116}, 2019.

\bibitem{infoxlm}
Chi, Z., L.~Dong, F.~Wei, et~al.
\newblock Infoxlm: An information-theoretic framework for cross-lingual
  language model pre-training.
\newblock \emph{arXiv preprint arXiv:2007.07834}, 2020.

\bibitem{UN}
Ziemski, M., M.~Junczys-Dowmunt, B.~Pouliquen.
\newblock The united nations parallel corpus v1. 0.
\newblock In \emph{Proceedings of the Tenth International Conference on
  Language Resources and Evaluation (LREC'16)}. 2016.

\bibitem{Bombay}
Kunchukuttan, A., P.~Mehta, P.~Bhattacharyya.
\newblock The iit bombay english-hindi parallel corpus.
\newblock \emph{arXiv preprint arXiv:1710.02855}, 2017.

\bibitem{OPUS}
Tiedemann, J.
\newblock Parallel data, tools and interfaces in opus.
\newblock In \emph{Lrec}. Citeseer, 2012.

\bibitem{WM}
Schwenk, H., V.~Chaudhary, S.~Sun, et~al.
\newblock Wikimatrix: Mining 135m parallel sentences in 1620 language pairs
  from wikipedia.
\newblock \emph{arXiv preprint arXiv:1907.05791}, 2019.

\bibitem{gelu}
Hendrycks, D., K.~Gimpel.
\newblock Gaussian error linear units (gelus).
\newblock \emph{arXiv preprint arXiv:1606.08415}, 2016.

\bibitem{Adam}
Kingma, D.~P., J.~Ba.
\newblock Adam: A method for stochastic optimization.
\newblock \emph{arXiv preprint arXiv:1412.6980}, 2014.

\bibitem{XNLI}
Conneau, A., G.~Lample, R.~Rinott, et~al.
\newblock Xnli: Evaluating cross-lingual sentence representations.
\newblock \emph{arXiv preprint arXiv:1809.05053}, 2018.

\bibitem{XLM}
Lample, G., A.~Conneau.
\newblock Cross-lingual language model pretraining.
\newblock \emph{arXiv preprint arXiv:1901.07291}, 2019.

\bibitem{Unicoder}
Huang, H., Y.~Liang, N.~Duan, et~al.
\newblock Unicoder: A universal language encoder by pre-training with multiple
  cross-lingual tasks.
\newblock \emph{arXiv preprint arXiv:1909.00964}, 2019.

\bibitem{VECO}
Luo, F., W.~Wang, J.~Liu, et~al.
\newblock Veco: Variable and flexible cross-lingual pre-training for language
  understanding and generation.
\newblock \emph{arXiv preprint arXiv:2010.16046}, 2020.

\bibitem{Label-smoothing}
Szegedy, C., V.~Vanhoucke, S.~Ioffe, et~al.
\newblock Rethinking the inception architecture for computer vision.
\newblock In \emph{Proceedings of the IEEE conference on computer vision and
  pattern recognition}. 2016.

\bibitem{Dropout}
Srivastava, N., G.~Hinton, A.~Krizhevsky, et~al.
\newblock Dropout: a simple way to prevent neural networks from overfitting.
\newblock \emph{The journal of machine learning research}, 2014.

\bibitem{straggler}
Dai, W., Y.~Zhou, N.~Dong, et~al.
\newblock Toward understanding the impact of staleness in distributed machine
  learning.
\newblock \emph{arXiv preprint arXiv:1810.03264}, 2018.

\bibitem{End-to-End}
Ao, Y., Z.~Wu, D.~Yu, et~al.
\newblock End-to-end adaptive distributed training on paddlepaddle.
\newblock \emph{arXiv preprint arXiv:2112.02752}, 2021.

\end{thebibliography}

\end{document}